\documentclass[11pt]{onesearchvl}

\usepackage{amsmath,amsfonts,bm}

\def\eqref#1{equation~\ref{#1}}

\def\1{\bm{1}}

\DeclareMathAlphabet{\mathsfit}{\encodingdefault}{\sfdefault}{m}{sl}
\SetMathAlphabet{\mathsfit}{bold}{\encodingdefault}{\sfdefault}{bx}{n}

\usepackage{wrapfig}
\usepackage{fvextra}
\usepackage{tabularx}
\usepackage{array}
\usepackage{xurl}
\usepackage{float}
\usepackage{needspace}
\usepackage{pifont}

\let\chapter\section

\fancypagestyle{firststyle}{\fancyhf{}\fancyfoot[C]{\small\thepage}}
\title{\raisebox{-0.16em}{\includegraphics[height=1.15em]{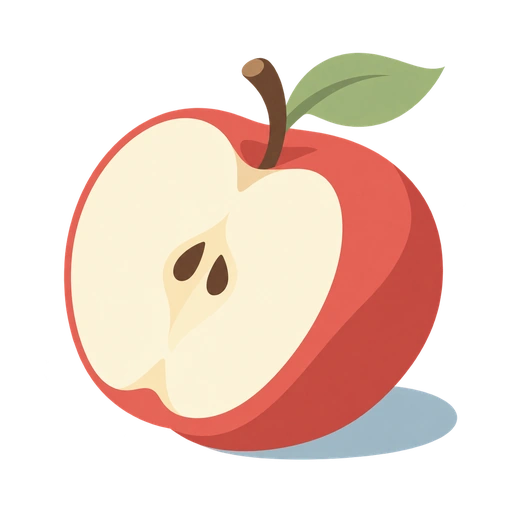}}\hspace{0.3em}PhysInsert: Physically Plausible Object Insertion with\\
Geometric Priors and Bidirectional Object-Scene Interactions}
\author[1,2,*]{Yue Gong}
\author[2]{Le Zhang}
\author[2]{Rui Chen}
\author[2]{Song Chang}
\author[2]{\\Xiaochao Qu}
\author[2]{Luoqi Liu}
\author[2,\dagger]{Ting Liu}
\author[3,\dagger]{Yebin Liu}
\affiliation[1]{School of Computer Science and Engineering, Beihang University\\}
\affiliation[2]{MT Lab, Meitu Inc.}
\affiliation[3]{Department of Automation, Tsinghua University}
\abstract{We present \textbf{PhysInsert}, a framework for physically plausible reference-based object insertion that jointly adapts objects and their surrounding scenes.
Existing methods often rely on masked inputs, generation-based pairs, or training pairs with limited viewpoint variation, leaving pose and illumination adaptation and bidirectional object--scene interactions insufficiently constrained.
We construct the \textbf{Insert World} dataset with strictly aligned Blender-rendered counterfactual pairs and object references across viewpoints and lighting conditions.
Building on these data, \emph{Geometry-Guided Adaptation} couples a VGGT geometry stream with image and text streams under target-scene feature supervision to guide spatially consistent insertion.
Complementary \emph{2D Object--Scene Spatial Supervision} jointly supervises object occupancy and environmental changes while grounding spatial instructions in object regions through image--text attention.
We introduce \textbf{PhysInsert-Bench} with dedicated cross-environment and cross-view cases requiring illumination and pose adaptation beyond appearance copying.
Experiments on PhysInsert-Bench and DreamBooth demonstrate improved image fidelity; A user study further show more accurate placement and improved lighting integration, perspective, and pose plausibility.
}
\checkdata[Project Page]{\url{https://arthuring.github.io/PhysInsert/}}
\hypersetup{
 pdftitle={PhysInsert: Physically Plausible Object Insertion with Geometric Priors and Bidirectional Object-Scene Interactions},
 pdfauthor={Yue Gong, Le Zhang, Rui Chen, Song Chang, Xiaochao Qu, Luoqi Liu, Ting Liu, Yebin Liu}
}
\begin{document}
\vspace*{-8mm}
\maketitle
\begingroup
\renewcommand{\thefootnote}{\fnsymbol{footnote}}
\footnotetext[1]{This work was conducted during internship at MT Lab, Meitu Inc.}
\endgroup
\begingroup
\renewcommand{\thefootnote}{\ensuremath{\dagger}}
\footnotetext[2]{Corresponding authors: \texttt{lt@meitu.com}; \texttt{liuyebin@tsinghua.edu.cn}.}
\endgroup
\begingroup
\setlength{\intextsep}{3pt}
\captionsetup{skip=5pt}
\begin{figure}[H]
    \centering
    \includegraphics[width=\linewidth]{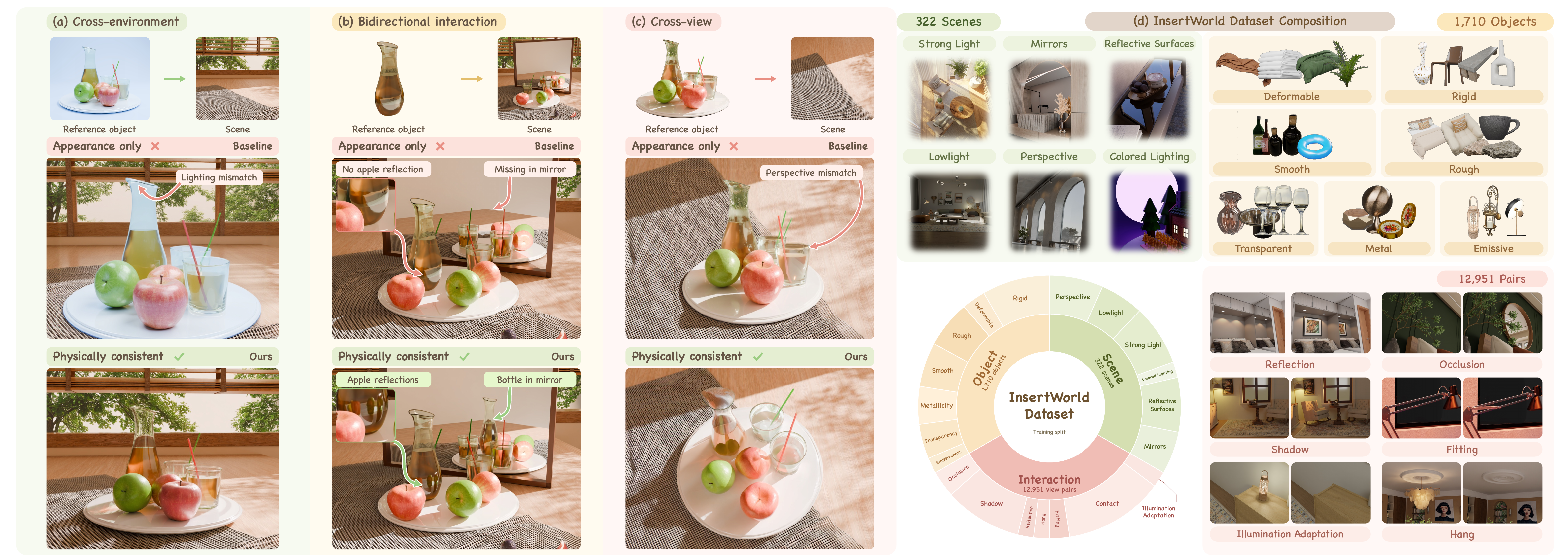}
    \caption{(a--c) PhysInsert simultaneously supports cross-environment adaptation, bidirectional object--scene interactions, and cross-view adaptation. (d) InsertWorld offers broad coverage of scenes, object properties, and interaction types.}
    \label{fig:motivation}
\end{figure}
\endgroup

\clearpage

\chapter{Introduction}

Reference-based object insertion aims to integrate a specified object into a target scene while preserving its identity.
Physically plausible insertion further requires geometric and optical consistency, with the object's perspective, pose, appearance, and effects on its surroundings all adapted to the target scene.

As illustrated in Figure~\ref{fig:motivation}, despite recent progress, three challenges remain in physically plausible object insertion.
\textbf{(a) Adaptation across environments.} Existing methods~\citep{Anydoor,InsertAnything,MIGE} can insert objects into new scenes, but provide limited controlled supervision for adapting the same object across lighting environments.
\textbf{(b) Bidirectional object--scene interactions.} Mask-and-fill methods~\citep{Direct,Ominipaint} may adapt object illumination yet struggle to produce consistent scene changes; training pairs constructed through masking or generative editing~\citep{Anydoor,InsertAnything} do not guarantee accurate supervision of shadows and reflections.
Training examples with simplified backgrounds~\citep{Object3DIT} also underrepresent complex occlusion, attachment, and spatial relationships with existing scene objects.
\textbf{(c) Adaptation across viewpoints.} Even with consistent illumination, differing reference and target viewpoints still require perspective and pose adaptation; fixed-view training pairs~\citep{PaintByExample,PSDiffusion} offer limited supervision for this change.
Reliance on mask or box guidance also limits fine-grained text control of placement and pose~\citep{Anydoor,Direct}.
Methods relying primarily on 2D image supervision~\citep{Anydoor,InsertAnything} lack direct constraints on object--scene geometry, making it difficult to guide adaptation to the target perspective and pose.

We propose \textbf{PhysInsert}, a framework combining counterfactual scene data, geometric priors, and spatial guidance for physically plausible object insertion.

To provide counterfactual supervision for object--scene adaptation, we construct \textbf{InsertWorld} from strictly aligned pairs rendered by changing only object presence in complete scenes, capturing both scene-dependent object appearance and object-induced environmental changes.
Object references rendered across lighting environments and viewpoints support cross-environment and cross-view adaptation.
Paired with spatial instructions, these examples support text-guided control of placement and complex interactions with existing scene objects.

To align object perspective and pose with the scene's 3D structure, \emph{Geometry-Guided Adaptation} models post-insertion geometry as a supervised intermediate representation.
Initialized from background VGGT features, a dedicated geometry stream evolves through joint attention with image and text tokens, with its feature readouts supervised against VGGT features of the with-object scene.
The geometry stream remains active at inference to guide scene-consistent object adaptation.

To localize insertion-induced changes and ground spatial instructions in the image, we complement geometric adaptation with two forms of 2D supervision.
\emph{Counterfactual Object--Scene Supervision} uses object masks and aligned scene pairs to jointly supervise object occupancy and external effects.
\emph{Soft-Mask Spatial Grounding} aligns image--text attention with the object region to translate spatial instructions into placement control while retaining the full background input.

To evaluate scene adaptation beyond appearance copying, we construct \textbf{PhysInsert-Bench} with cases requiring illumination or pose adaptation and include text-guided and real-world evaluations.

Our main contributions are as follows:
\begin{itemize}
    \item \textbf{Dataset:} InsertWorld combines strictly aligned counterfactual scene pairs, cross-view and cross-lighting object references, and spatial instructions to support learning object adaptation and complex object--scene interactions.
    \item \textbf{Method:} PhysInsert combines a background-initialized geometry stream supervised toward post-insertion scene features with 2D supervision, improving cross-environment and cross-view insertion and text-guided placement.
    \item \textbf{Benchmark:} PhysInsert-Bench tests illumination and pose adaptation beyond appearance copying through dedicated cross-environment, cross-view, and in-the-wild cases.
\end{itemize}

\chapter{Related Work}
\label{sec:related-work}

\paragraph{Physics-aware Image Generation and Editing.}
Recent work has sought to improve the physical plausibility of image editing by modeling scene geometry and physical interactions.
PhysicEdit~\citep{PhysicEdit} learns video-based transition priors to model causal physical changes in instruction-based editing.
PhyEdit~\citep{PhyEdit} relocates existing objects within a scene using geometric guidance and depth supervision.
Object3DIT~\citep{Object3DIT} supports language-guided translation, rotation, insertion, and removal in procedurally rendered scenes; its insertion specifies objects through text rather than external reference images.
Our task requires preserving a reference object's identity while adapting its viewpoint, illumination, and interactions to a new scene.

\paragraph{Reference-based Object Insertion.}
For supervision, ObjectDrop~\citep{Objectdrop} and OmniPaint~\citep{Ominipaint} use real counterfactual pairs, supplemented by generated bootstrapping or insertion--removal cycles.
ObjectMate~\citep{ObjectMate} and Insert-Anything~\citep{InsertAnything} construct backgrounds through learned editing, whose accuracy depends on the editor; PS-Diffusion~\citep{PSDiffusion} uses rendered pairs.
For control, mask-based methods~\citep{PaintByExample,Anydoor,Ominipaint} discard background information within the editing region, while ObjectMate uses boxes, Insert-Anything also supports text guidance, and DIRECT~\citep{Direct} requires 3D proxy reconstruction and placement for pose control.
For viewpoint adaptation, Paint by Example derives same-image references, PS-Diffusion retains the target viewpoint, and ObjectDrop preserves pasted-object pose and illumination; AnyDoor, Insert-Anything, and ObjectMate incorporate multiview data.
PhysInsert combines controlled counterfactual rendering and diverse references with supervised geometry states, object--effect supervision, and spatial-text grounding.

\chapter{InsertWorld Dataset}

This section compares InsertWorld with existing datasets, describes its construction pipeline, and summarizes its statistics of training--test split.

\par\addvspace{\baselineskip}
\noindent\begin{minipage}{\linewidth}
\subsection{Comparison with Existing Datasets}
\label{sec:dataset-comparison}
\noindent\begin{minipage}[t]{0.46\linewidth}
\vspace{0pt}
\captionsetup{justification=raggedright,singlelinecheck=false}
\captionof{table}{Comparison of object-insertion datasets.}
\label{tab:dataset-comparison}
\begingroup
\centering
\setlength{\tabcolsep}{2.5pt}
\renewcommand{\arraystretch}{1.2}
\footnotesize
\newcommand{\datasetyes}{\textcolor{onesearchpurple}{\ding{51}}}
\newcommand{\datasetno}{\textcolor{red!75!black}{\ding{55}}}
\begin{tabular*}{\linewidth}{@{\extracolsep{\fill}}lcccc@{}}
\toprule
Dataset & CF & MV & ML & Ctrl. \\
\midrule
AnyInsertion & \datasetno & Partial & \datasetno & Mask / Text \\
DIRECT & \datasetno & \datasetyes & \datasetno & Mask + pose \\
\midrule
InsertWorld & \datasetyes & \datasetyes & \datasetyes & Text / Box \\
\bottomrule
\end{tabular*}
\par
\vspace{4pt}
\raggedright\scriptsize
CF: controlled counterfactual pairs; MV: multiview references; ML: multilight references; Ctrl.: control inputs.
\datasetyes: provided; \datasetno: not specifically constructed; Partial: partial coverage.
\par\endgroup

\end{minipage}\hfill
\begin{minipage}[t]{0.50\linewidth}
\vspace{0pt}
As summarized in Table~\ref{tab:dataset-comparison}, AnyInsertion~\citep{InsertAnything} and DIRECT~\citep{Direct} do not explicitly provide controlled counterfactual scene pairs or dedicated multilight references. Multiview references cover only part of AnyInsertion. Our InsertWorld combines strictly aligned counterfactual pairs, multiview and multilight references to support adaptation across environments and viewpoints alongside object--scene interactions.
\end{minipage}
\end{minipage}
\par\addvspace{0.4\baselineskip}

\subsection{Dataset Construction}
\label{sec:dataset-construction}

\begin{figure*}[t]
    \centering
    \includegraphics[width=\linewidth]{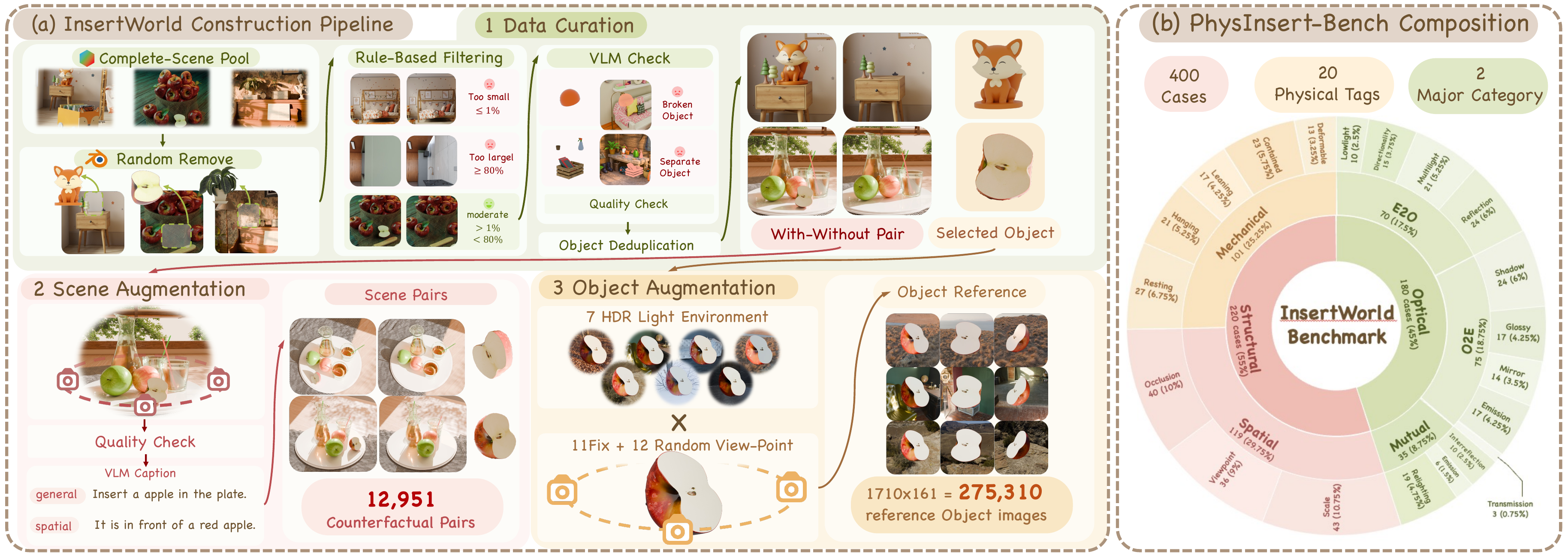}
    \caption{Overview of InsertWorld construction and PhysInsert-Bench composition.}
    \label{fig:dataset-construction}
\end{figure*}

To reduce interpenetration and floating artifacts from random placement, we construct InsertWorld by removing existing objects from complete BlenderKit scenes (Figure~\ref{fig:dataset-construction}). We render each scene with and without the object under fixed camera, lighting, and remaining scene settings, recomputing its shadows, mirror images, and reflections. These strictly aligned counterfactual pairs capture scene-dependent object appearance and object-induced environmental changes. Quality filtering and object deduplication retain suitable scene--object combinations.

We augment scene viewpoints and independently render each retained object across lighting environments and viewpoints to provide references for cross-environment and cross-view adaptation. Each example combines an object-free background, a with-object target, and reference images with Gemini-generated annotations comprising an editing instruction, a placement description, and a target-pose description. Filtering and sampling details are provided in Appendix~\ref{app:dataset-construction}.

\subsection{Dataset Overview}
\label{sec:dataset-overview}

The InsertWorld training set contains 12,951 counterfactual scene image pairs and 275,310 reference images of 1,710 objects spanning different viewpoints and lighting conditions. For evaluation, PhysInsert-Bench comprises 400 cases from 103 scenes with no overlap with the training scenes and object, including 180 \textit{Optical} cases (154 with aligned reference--target viewpoints and 26 cross-view challenges) and 220 \textit{Structural} cases with independent reference viewpoints. Benchmark construction is detailed in \S\ref{sec:benchmark-construction}.

\chapter{PhysInsert Model}

This section introduces our geometry-conditioned object insertion model and then describes the construction of PhysInsert-Bench.

Given an object-free background $I_{-}$, reference-object images $\mathcal{I}_r=\{I_r^i\}_{i=1}^{N_r}$, and an instruction $p$, we aim to generate the with-object image $I_{+}$ while preserving object identity and adapting both the object and its environmental effects to the target scene.

\subsection{Geometry-Conditioned Object Insertion}

\begin{figure*}[t]
    \centering
    \includegraphics[width=\linewidth]{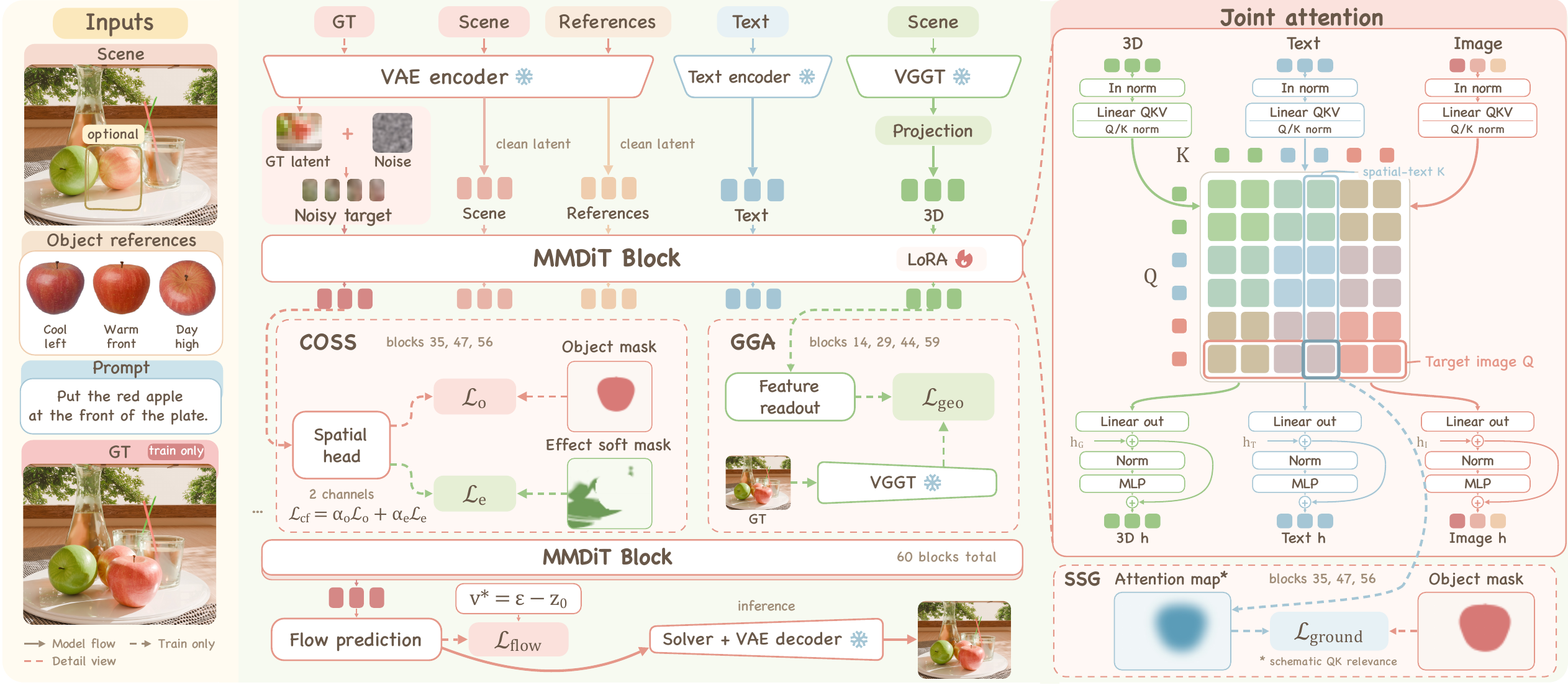}
    \caption{PhysInsert combines a VGGT geometry stream with image and text streams. Target-scene features supervise geometry-token states, while 2D-OSS aligns object/effect predictions and spatial-text attention with image regions. Auxiliary supervision is used only during training.}
    \label{fig:method-overview}
\end{figure*}

Our method combines Geometry-Guided Adaptation (GGA) for scene-consistent object adaptation with 2D Object--Scene Spatial Supervision (2D-OSS) for object regions, environmental changes, and spatial-text alignment (Figure~\ref{fig:method-overview}). In our Qwen-Image-Edit-2511 backbone, the image stream $U_0$ contains the noisy target latent and frozen-VAE encodings of the complete background and references; a frozen multimodal encoder produces text tokens $T_0$ from the instruction and image context. Implementation and loss details are provided in Appendix~\ref{app:method-details}.

\subsubsection{Geometry-Guided Adaptation}

To guide scene-consistent object adaptation, GGA initializes a dynamic geometry stream from background features $F_{-}$ extracted by frozen VGGT~\citep{VGGT}. An MLP $A_\phi$ maps these features to the DiT hidden space, with a learned geometry-modality embedding $e_G$. Geometry, image, and text states interact through $L$ three-stream DiT blocks:
\begin{equation}
\begin{aligned}
    G_0 &= A_\phi(F_{-})+e_G,\\
    (U_{\ell+1},T_{\ell+1},G_{\ell+1})
    &=\mathcal{B}_\ell(U_\ell,T_\ell,G_\ell),\qquad \ell=0,\ldots,L-1.
\end{aligned}
\label{eq:geometry-stream}
\end{equation}
At each block, the three streams compute their own queries, keys, and values, concatenate them for joint attention, and update through stream-specific modulation, output projections, and feed-forward paths. The geometry states thus evolve through interaction with the target-image, background, reference, and text conditions throughout generation.

Background features describe the scene before insertion and alone do not specify its geometry after adding the object. We therefore extract target features $F_{+}$ using the same frozen VGGT and train a shared readout $P_\psi$ to predict them from geometry states at selected layers $\mathcal{S}_{\mathrm{geo}}$. This supervision guides the dynamic geometry stream toward the target scene structure, complementing the initial background prior. The Geometry Alignment Loss is
\begin{equation}
\begin{aligned}
    \mathcal{L}_{\mathrm{geo}}
    &=\sum_{\ell\in\mathcal{S}_{\mathrm{geo}}}\omega_\ell
    \mathcal{D}_{\mathrm{geo}}\!\left(P_\psi(G_\ell),F_{+}\right),\\
    \mathcal{D}_{\mathrm{geo}}(A,B)
    &=\operatorname{SL1}_{\beta_{\mathrm{geo}}}\!\left(A,\operatorname{LN}(B)\right)
    +\lambda_{\mathrm{cos}}\operatorname{CosDist}\!\left(A,\operatorname{LN}(B)\right).
\end{aligned}
\label{eq:geometry-alignment}
\end{equation}
Here $A$ and $B$ denote predicted and target features, $\operatorname{LN}$ is layer normalization, $\beta_{\mathrm{geo}}$ controls the Smooth L1 transition, and $\lambda_{\mathrm{cos}}$ weights cosine distance; $\omega_\ell$ are normalized nonnegative layer weights. Target features are stop-gradient supervision only and are never injected into the generator.

\subsubsection{2D Object--Scene Spatial Supervision}

Complementing GGA's geometric feature supervision, 2D-OSS directly constrains where the inserted object and its visible environmental effects should appear in the image, while aligning spatial instructions with the object region.

\paragraph{Counterfactual Object--Scene Supervision (COSS).}
We downsample the ground-truth object mask to the target-token grid to obtain a soft object occupancy map $M_o$. From the aligned counterfactual pair, we compute a smoothed difference map $D$ and exclude the object region to construct the external-effect map $M_e$:
\begin{equation}
\begin{aligned}
    D &= \operatorname{AvgPool}_{\kappa\times\kappa}\!\left(
    \operatorname{Mean}_{\mathrm{RGB}}
    \left|\operatorname{Down}(I_{+})-\operatorname{Down}(I_{-})\right|
    \right),\\
    M_e &= \operatorname{clip}\!\left(
    \frac{D-\delta_{\mathrm{low}}}{\delta_{\mathrm{high}}-\delta_{\mathrm{low}}},0,1\right)
    \left[1-\operatorname{Dilate}_{\rho_d}\!\left(\mathbf{1}[M_o>\delta_{\mathrm{obj}}]\right)\right].
\end{aligned}
\label{eq:counterfactual-targets}
\end{equation}
Here $\operatorname{Down}$ maps images to the target-token grid, $\operatorname{Mean}_{\mathrm{RGB}}$ averages channels, and $\kappa$ controls smoothing. Thresholds $\delta_{\mathrm{low}}<\delta_{\mathrm{high}}$ map differences to soft effect strengths; $\delta_{\mathrm{obj}}$ and dilation radius $\rho_d$ exclude the object and its boundary neighborhood. Thus, $M_e$ provides proxy labels for visible changes such as shadows and reflections, rather than a semantic decomposition of optical effects.

A shared spatial head operates on target-image tokens at layers $\mathcal{S}_{\mathrm{cf}}$, averaging its logits across layers before producing object and effect probability maps. The Object Occupancy Loss $\mathcal{L}_o$ and Scene Effect Loss $\mathcal{L}_e$ each combine balanced binary cross-entropy and Dice losses, forming the Counterfactual Supervision Loss:
\begin{equation}
    \mathcal{L}_{\mathrm{cf}}=\alpha_o\mathcal{L}_o+\alpha_e\mathcal{L}_e,
    \qquad \alpha_o,\alpha_e\geq0,\quad \alpha_o+\alpha_e=1.
\label{eq:counterfactual-supervision}
\end{equation}

\paragraph{Soft-Mask Spatial Grounding (SSG).}
SSG grounds spatial instructions in the object region. The spatial description is a dedicated field of the generated caption (Figure~\ref{fig:dataset-construction}); we use this field to identify its corresponding text-token subset $\mathcal{T}_s$ in the instruction. At layer $\ell$ and head $h$, let $q_i^{(\ell,h)}$ denote the query at target-image position $i$ and $k_j^{(\ell,h)}$ the key of spatial-text token $j$. Their relevance is aggregated over phrase tokens using log-mean-exp and then averaged over heads:
\begin{equation}
    r_i^{(\ell)}
    =\frac{1}{N_h}\sum_{h=1}^{N_h}
    \log\!\left[
    \frac{1}{|\mathcal{T}_s|}
    \sum_{j\in\mathcal{T}_s}
    \exp\!\left(\frac{q_i^{(\ell,h)}\cdot k_j^{(\ell,h)}}{\sqrt{d_h}}\right)
    \right],
\label{eq:spatial-relevance}
\end{equation}
where $N_h$ is the number of heads and $d_h$ the query/key dimension. With temperature $\tau_a>0$, we normalize relevance over target-image positions and the soft object map over the same grid:
\begin{equation}
    P_r^{(\ell)}(i)=\operatorname{softmax}_i(r_i^{(\ell)}/\tau_a),
    \qquad
    P_o(i)=\frac{M_o(i)}{\sum_{i'}M_o(i')}.
\label{eq:grounding-distributions}
\end{equation}
The Object Grounding Loss aligns these distributions over the supervised layer set $\mathcal{S}_{\mathrm{ground}}$:
\begin{equation}
    \mathcal{L}_{\mathrm{ground}}=
    \frac{1}{|\mathcal{S}_{\mathrm{ground}}|}
    \sum_{\ell\in\mathcal{S}_{\mathrm{ground}}}
    D_{\mathrm{KL}}\!\left(P_o\,\middle\|\,P_r^{(\ell)}\right).
\label{eq:object-grounding}
\end{equation}
Together with COSS, which supervises object occupancy and external effects, this alignment connects textual placement requirements to the region occupied by the inserted object.

\paragraph{Overall Objective.}
We combine COSS and SSG as the 2D Object--Scene Spatial Supervision Loss and optimize
\begin{equation}
\begin{aligned}
    \mathcal{L}_{\mathrm{2D}}
    &=\lambda_{\mathrm{cf}}\mathcal{L}_{\mathrm{cf}}
    +\lambda_{\mathrm{ground}}\mathcal{L}_{\mathrm{ground}},\\
    \mathcal{L}
    &=\mathcal{L}_{\mathrm{flow}}
    +\lambda_{\mathrm{geo}}\mathcal{L}_{\mathrm{geo}}
    +\mathcal{L}_{\mathrm{2D}}.
\end{aligned}
\label{eq:overall-objective}
\end{equation}
with nonnegative weights $\lambda_{\mathrm{geo}}$, $\lambda_{\mathrm{cf}}$, and $\lambda_{\mathrm{ground}}$. Text-guided inference uses $I_{-}$, $\mathcal{I}_r$, and $p$. For box-guided inference, we additionally supply a reference image annotated with the target bounding box as visual conditioning. All three streams remain active in both settings; target images, object masks, and auxiliary heads are used only during training.

\subsection{Benchmark Construction}
\label{sec:benchmark-construction}

As shown in Figure~\ref{fig:dataset-construction}(b), PhysInsert-Bench evaluates optical and spatial--pose adaptation using aligned counterfactual scene pairs. \textit{Optical} cases emphasize illumination differences and visible object-induced scene changes; their challenge cases and \textit{Structural} cases additionally provide independent reference viewpoints to test pose adaptation. Each case is annotated for bidirectional interactions in lighting, spatial relationships, and pose, with physical-effect labels and evaluation-focus tags identifying the effects emphasized by that case.

\chapter{Experiments}

\begin{table*}[t]
    \centering
    \caption{Text-guided results on PhysInsert-Bench using one object-reference image per case for all reported metrics. Phy-VLM obtains three valid ratings per output with GPT-5.5, retrying null responses, and reports mean $\pm$ 95\% confidence-interval half-width. Unless otherwise specified, the following conventions apply to this and subsequent tables: +PhysW. denotes fine-tuning on our InsertWorld dataset with the same training budget as ours under the corresponding guidance setting; \textbf{bold}/\underline{underline} mark the best/second-best across methods.}
    \label{tab:text-guide}
    \definecolor{proprietaryblue}{RGB}{224,234,247}
    \definecolor{opensourcepink}{RGB}{253,224,224}
    \setlength{\tabcolsep}{4.0pt}
    \renewcommand{\arraystretch}{1.08}
    \scriptsize
    \resizebox{\textwidth}{!}{%
    \begin{tabular}{lcccccccccc}
        \toprule
        \textbf{Method}
        & DIoU (\%)$\uparrow$ & Mask IoU (\%)$\uparrow$ & DINO-Obj.$\uparrow$ & PSNR$\uparrow$
        & SSIM$\uparrow$ & LPIPS$\downarrow$ & FID$\downarrow$
        & AbsRel$\downarrow$ & $\delta_{1.25}$ (\%)$\uparrow$
        & \shortstack{Phy-VLM$\uparrow$} \\
        \midrule
        \rowcolor{proprietaryblue}
        \multicolumn{11}{c}{\textit{\textbf{Proprietary Models}}} \\
        \midrule
        Nano Banana Pro~\citep{NanoBananaPro}
        & \underline{54.58} & \underline{48.34} & \underline{0.7023} & \underline{22.85}
        & 0.8231 & \underline{0.1527} & \underline{46.55}
        & \underline{0.2066} & \underline{77.08}  & $\underline{3.48} \pm 0.09$ \\
        GPT-Image-2~\citep{GPTImage2}
        & 38.02 & 31.92 & 0.6746 & 19.39
        & 0.7017 & 0.2949 & 53.46
        & 0.3129 & 65.48  & $3.17 \pm 0.11$ \\
        \midrule
        \rowcolor{opensourcepink}
        \multicolumn{11}{c}{\textit{\textbf{Open-Source Models}}} \\
        \midrule
        OmniPaint + PhysW.~\citep{Ominipaint}
        & 37.69 & 33.94 & 0.5828 & 20.67
        & \underline{0.8302} & 0.1733 & 73.54
        & 0.2401 & 67.30  & $2.06 \pm 0.13$ \\
        Insert-Anything + PhysW.~\citep{InsertAnything}
        & 22.66 & 21.46 & 0.6275 & 19.45
        & 0.7711 & 0.3128 & 64.39
        & 0.2651 & 67.79  & $2.36 \pm 0.13$ \\
        JoyAI-Image-Edit-Plus~\citep{JoyAIImage}
        & 46.37 & 39.21 & 0.6614 & 21.13
        & 0.8211 & 0.1769 & 56.37
        & 0.2308 & 75.70  & $3.04 \pm 0.09$ \\
        \midrule
        Qwen-Image-Edit-2511~\citep{QwenImage}
        & 48.32 & 44.14 & 0.6786 & 19.34
        & 0.7840 & 0.2260 & 60.99
        & 0.2207 & 73.35  & $2.94 \pm 0.15$ \\
        \textbf{Ours}
        & \textbf{62.99} & \textbf{56.08} & \textbf{0.7276} & \textbf{24.40}
        & \textbf{0.8773} & \textbf{0.1118} & \textbf{41.96}
        & \textbf{0.1195} & \textbf{86.56}  & $\mathbf{3.52} \pm 0.11$ \\
        \bottomrule
    \end{tabular}%
    }
\end{table*}

\begin{table*}[t]
    \centering
    \caption{Box-guided results on PhysInsert-Bench. Ours is adapted from the text-guided model for an additional 4,000 box-guided steps; +PhysW. baselines use the same training budget with our method. $\dagger$ For PhysInsert-Bench, DIRECT uses GT-image-assisted pose fitting}
    \label{tab:box-guide}
    \definecolor{proprietaryblue}{RGB}{224,234,247}
    \definecolor{opensourcepink}{RGB}{253,224,224}
    \setlength{\tabcolsep}{4.0pt}
    \renewcommand{\arraystretch}{1.08}
    \scriptsize
    \resizebox{\textwidth}{!}{%
    \begin{tabular}{lcccccccc}
        \toprule
        \textbf{Method}
        & Mask IoU (\%)$\uparrow$ & DINO-Obj.$\uparrow$ & PSNR$\uparrow$
        & SSIM$\uparrow$ & LPIPS$\downarrow$ & FID$\downarrow$
        & AbsRel$\downarrow$ & $\delta_{1.25}$ (\%)$\uparrow$ \\
        \midrule
        \rowcolor{proprietaryblue}
        \multicolumn{9}{c}{\textit{\textbf{Proprietary Models}}} \\
        \midrule
        Nano Banana Pro~\citep{NanoBananaPro}
        & 52.94 & \underline{0.7152} & 22.53 & 0.8056 & 0.1609 & 45.32 & 0.1846 & 81.03 \\
        GPT-Image-2~\citep{GPTImage2}
        & 39.61 & 0.6798 & 21.45 & 0.7716 & 0.2049 & 50.68 & 0.2860 & 71.24 \\
        \midrule
        \rowcolor{opensourcepink}
        \multicolumn{9}{c}{\textit{\textbf{Open-Source Models}}} \\
        \midrule
        AnyDoor~\citep{Anydoor}
        & 55.92 & 0.5379 & 20.42 & 0.8158 & 0.2413 & 63.50 & 0.2070 & 67.01 \\
        Insert-Anything~\citep{InsertAnything}
        & 55.09 & 0.6507 & 22.70 & 0.8735 & 0.1608 & 52.23 & 0.1911 & 74.99 \\
        
        OmniPaint~\citep{Ominipaint}
        & 40.72 & 0.5441 & 20.60 & 0.8473 & 0.1402 & 59.95 & 0.2285 & 72.55 \\
        
        FreeCompose~\citep{FreeCompose}
        & 55.62 & 0.5037 & 20.85 & 0.7666 & 0.3288 & 62.62 & 0.2497 & 62.70 \\
        PS-Diffusion~\citep{PSDiffusion}
        & 60.58 & 0.5098 & 20.41 & 0.7384 & 0.2511 & 60.35 & 0.2122 & 68.49 \\
        SHINE-Adapter~\citep{SHINE}
        & 52.07 & 0.5567 & 21.94 & 0.7910 & 0.1798 & 65.77 & 0.1976 & 78.18 \\
        DIRECT + GT Pose$\dagger$~\citep{Direct}
        & \underline{65.86} & 0.6626 & 23.14 & \textbf{0.8957} & \underline{0.1270} & 49.97 & 0.2180 & 68.05 \\
        Insert-Anything + PhysW.~\citep{InsertAnything}
        & 60.86 & 0.6991 & \underline{24.38} & 0.8759 & 0.1554 & \underline{43.80} & \underline{0.1258} & \underline{86.56} \\
        OmniPaint + PhysW.~\citep{Ominipaint}
        & 54.43 & 0.5447 & 22.49 & 0.8505 & 0.1290 & 57.85 & 0.1483 & 80.47 \\
        \midrule
        Qwen-Image-Edit-2511~\citep{QwenImage}
        & 37.39 & 0.5366 & 21.27 & 0.8605 & 0.1601 & 65.27 & 0.2654 & 71.85 \\
        \textbf{Ours}
        & \textbf{68.62} & \textbf{0.7229} & \textbf{25.81} & \underline{0.8876} & \textbf{0.1047} & \textbf{37.50} & \textbf{0.1064} & \textbf{89.41} \\
        \bottomrule
    \end{tabular}%
    }
\end{table*}

\begin{table*}[t]
    \centering
    \caption{Results on PhysInsert-Bench-Wild with one or three reference images, and DreamBooth with one reference image. }
    \label{tab:bench_wild}
    \definecolor{proprietaryblue}{RGB}{224,234,247}
    \definecolor{opensourcepink}{RGB}{253,224,224}
    \setlength{\tabcolsep}{3.5pt}
    \renewcommand{\arraystretch}{1.08}
    \scriptsize
    \resizebox{\textwidth}{!}{%
    \begin{tabular}{lccccccccc}
        \toprule
        \multirow{3}{*}{\textbf{Method}}
        & \multicolumn{6}{c}{\textbf{PhysInsert-Bench-Wild}}
        & \multicolumn{3}{c}{\textbf{DreamBooth}} \\
        \cmidrule(lr){2-7}\cmidrule(lr){8-10}
        & \multicolumn{3}{c}{\textbf{1-view}}
        & \multicolumn{3}{c}{\textbf{3-view}}
        & \multicolumn{3}{c}{\textbf{1-view}} \\
        \cmidrule(lr){2-4}\cmidrule(lr){5-7}\cmidrule(lr){8-10}
        & PSNR$\uparrow$ & FID$\downarrow$ & DINO-Obj.$\uparrow$
        & PSNR$\uparrow$ & FID$\downarrow$ & DINO-Obj.$\uparrow$
        & PSNR$\uparrow$ & FID$\downarrow$ & DINO-Obj.$\uparrow$ \\
        \midrule
        \rowcolor{proprietaryblue}
        \multicolumn{10}{c}{\textit{\textbf{Proprietary Models}}} \\
        \midrule
        Nano Banana Pro~\citep{NanoBananaPro}
        & 15.78 & 123.2 & \textbf{0.7881}
        & 15.72 & 115.3 & \textbf{0.7909}
        & 15.58 & \underline{83.74} & 0.8650 \\
        GPT-Image-2~\citep{GPTImage2}
        & 16.74 & \underline{113.4} & 0.7457
        & \underline{16.88} & \underline{112.9} & \underline{0.7865}
        & 14.68 & 87.77 & \underline{0.8821} \\
        \midrule
        \rowcolor{opensourcepink}
        \multicolumn{10}{c}{\textit{\textbf{Open-Source Models}}} \\
        \midrule
        OmniPaint~\citep{Ominipaint}
        & 18.08 & 140.5 & 0.6393
        & -- & -- & --
        & 15.10 & 115.8 & 0.7190 \\
        Insert-Anything~\citep{InsertAnything}
        & \underline{19.38} & 122.3 & 0.7536
        & -- & -- & --
        & \underline{16.95} & 90.53 & 0.8545 \\
        Qwen-Image-Edit-2511~\citep{QwenImage}
        & 12.87 & 208.1 & 0.5991
        & 10.93 & 214.6 & 0.5787
        & 16.58 & 86.17 & 0.8767 \\
        \textbf{Ours}
        & \textbf{20.36} & \textbf{95.67} & \underline{0.7784}
        & \textbf{20.60} & \textbf{87.90} & 0.7849
        & \textbf{17.45} & \textbf{83.68} & \textbf{0.8856} \\
        \bottomrule
    \end{tabular}%
    }
\end{table*}

\subsection{Experimental Setup}

\paragraph{Implementation Details.}
Our method builds upon Qwen-Image-Edit-2511, a DiT-based image editing model, retaining its pretrained Qwen2.5-VL~\citep{Qwen25VL} multimodal encoder and VAE and adding a VGGT geometry stream alongside the image and text streams. We fine-tune the DiT using LoRA~\citep{LoRA} with a rank of 32. Training is performed with a batch size of 8 and a maximum training-image pixel count of $1024^2$. We use the AdamW optimizer~\citep{AdamW} with a learning rate of $10^{-4}$ and a weight decay of $0.01$. All experiments are conducted on eight NVIDIA L20X GPUs. We train our model in a single stage on our proposed InsertWorld dataset for approximately 16,000 steps. At inference, we use 40 denoising steps and a CFG~\citep{ClassifierFreeGuidance} scale of 4.0. Full training and inference hyperparameters are provided in Appendix~\ref{app:hyperparameters}.

\paragraph{Test Datasets.}
We evaluate on PhysInsert-Bench (\S\ref{sec:benchmark-construction}), 40 DreamBooth~\citep{DreamBooth} cases covering all objects with one reference per case, and 50 PhysInsert-Bench-Wild cases with multiview reference pools. The Wild set draws on ABO~\citep{ABO}, NAVI v1.5~\citep{NAVI}, ORIDa~\citep{ORIDa}, and real target images paired with reference views rendered from SAM 3D Objects~\citep{SAM3DObjects} reconstructions. We favor reference poses that differ from the target pose to test cross-view adaptation; construction details are provided in Appendix~\ref{app:real-world-test-construction}.

\paragraph{Evaluation Metrics.}
We use PSNR and SSIM~\citep{SSIM}, LPIPS~\citep{LPIPS}, and FID~\citep{FID} to evaluate full-image fidelity, and DINO-Obj. to assess object identity preservation. Following PhyEdit~\citep{PhyEdit}, we use DIoU to measure object placement with a center-distance penalty, while Mask IoU measures visible-object overlap, including differences in visible extent caused by occlusion. AbsRel and $\delta_{1.25}$ assess estimated object-region depth consistency as a proxy for geometric plausibility. We also adapt the VLM-based physical plausibility evaluation of PhyEdit~\citep{PhyEdit} as Phy-VLM, using GPT-5.5 with a fixed prompt template augmented by each case's physical-effect annotations and evaluation-focus tags (\S\ref{sec:benchmark-construction}). These tags direct attention to the physical effects emphasized by that case. Metric definitions and implementation details are provided in Appendix~\ref{app:evaluation-metrics}.

\subsection{Quantitative Results}

\paragraph{Text-Guided Results.}
All metrics in Table~\ref{tab:text-guide} use one object-reference image per case. With placement specified through text instructions, PhysInsert achieves the highest DIoU and Mask IoU in Table~\ref{tab:text-guide}, indicating more accurate object placement. It also leads the reported fidelity and identity metrics and achieves the best estimated depth consistency under the symmetric DA3 protocol. In the Phy-VLM evaluation, PhysInsert obtains the highest mean score.

\paragraph{Box-Guided Results.}
With one reference image per case, Table~\ref{tab:box-guide} shows that PhysInsert achieves the highest DINO-Obj. and PSNR, together with the lowest LPIPS and FID. Fine-tuning on InsertWorld (+PhysW.) improves all reported metrics for both Insert-Anything and OmniPaint under the same training budget, showing that our data also benefits existing insertion models.

\paragraph{Real-World Results.}
On PhysInsert-Bench-Wild and DreamBooth (Table~\ref{tab:bench_wild}), PhysInsert achieves the highest PSNR and lowest FID across the reported reference-view settings. Its DINO-Obj. scores lead the open-source methods throughout and all compared methods on DreamBooth.

\begin{figure}[t]
    \centering
    \includegraphics[width=0.95\linewidth]{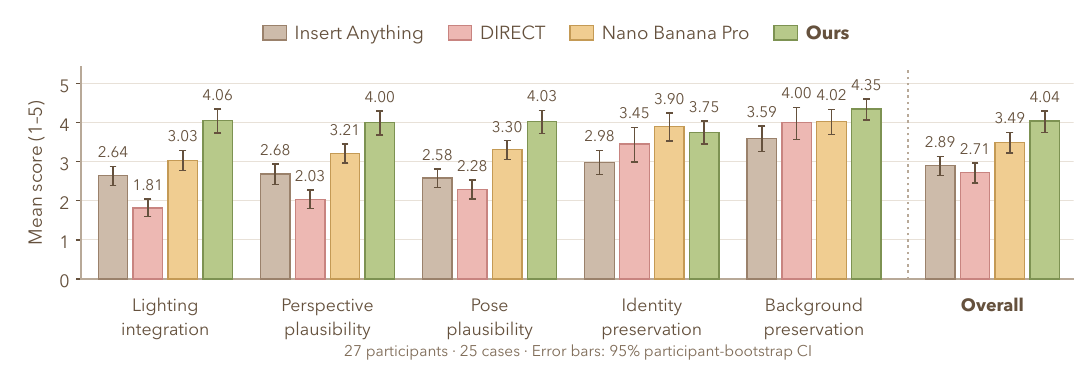}
    \caption{User-study result. Error bars show 95\% confidence intervals.}
    \label{fig:user-study}
\end{figure}

\paragraph{User Study.}
In a blinded study, 27 participants rate 25 separate cases combining RealBokeh backgrounds and Distinctions-646 objects (Appendix~\ref{app:user-study-protocol}). PhysInsert receives the highest ratings for lighting integration, perspective, and pose, and the highest overall score of 4.04, while remaining competitive in identity and background preservation (Figure~\ref{fig:user-study}).

\subsection{Qualitative Results}

\begin{figure*}[t]
    \centering
    \includegraphics[width=\linewidth]{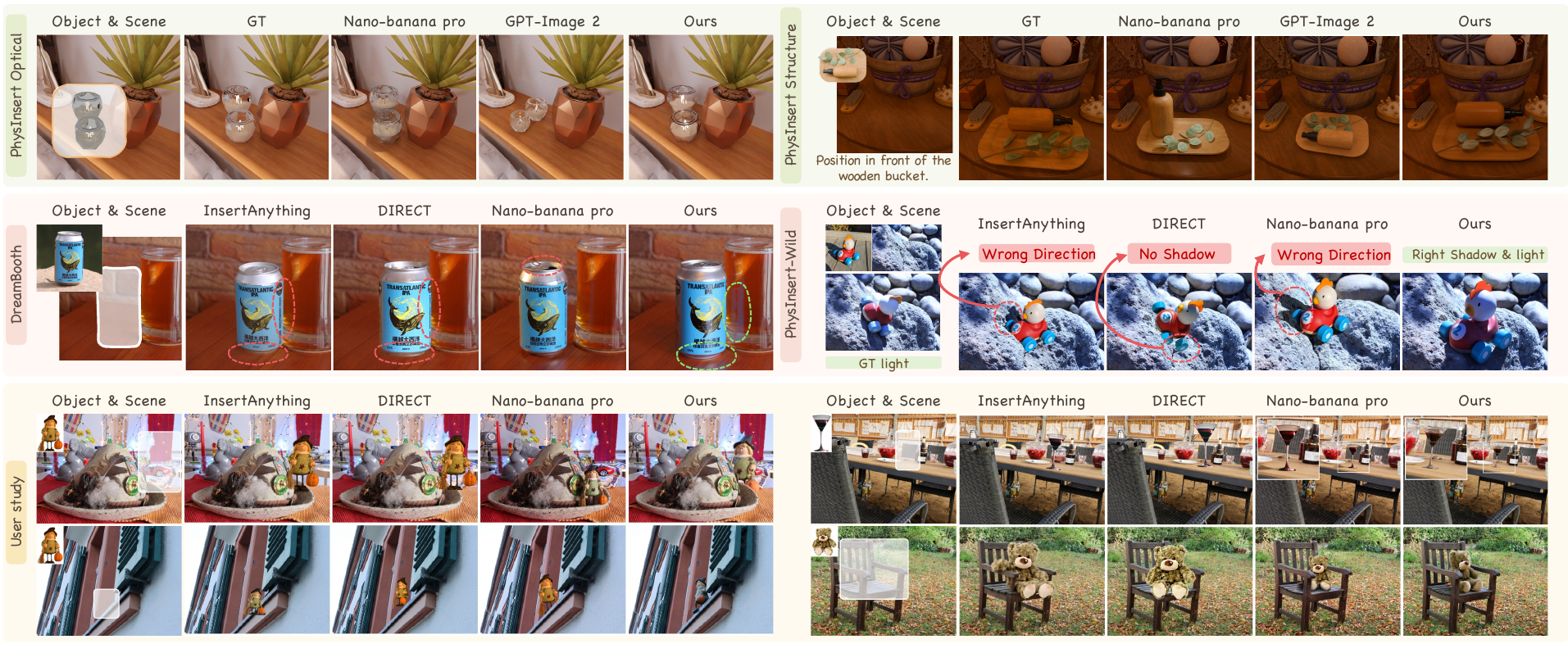}
    \caption{Qualitative comparisons on PhysInsert-Bench (top), DreamBooth and PhysInsert-Bench-Wild (middle), and the user-study set (bottom).}
    \label{fig:qualitative-comparison}
\end{figure*}

Figure~\ref{fig:qualitative-comparison} compares PhysInsert-Bench optical and structural adaptation cases (top), DreamBooth and PhysInsert-Bench-Wild cases (middle), and user-study examples (bottom). PhysInsert better adapts the reflective containers and beverage can to scene lighting, including the can's reflection in the adjacent glass. The tray and toy examples show orientation adaptation, with the toy's cast shadow also matching the target scene more closely. User-study examples further illustrate foreground occlusion, perspective adaptation on an oblique window ledge, and object scale on a chair.

\begingroup
\setlength{\parskip}{3pt plus 1pt}
\setlength{\textfloatsep}{8pt plus 2pt minus 2pt}
\setlength{\floatsep}{6pt plus 2pt minus 2pt}
\setlength{\intextsep}{8pt plus 2pt minus 2pt}
\captionsetup{skip=5pt}
\subsection{Ablation Studies}

\begin{figure}[!t]
    \centering
    \includegraphics[width=0.9\linewidth]{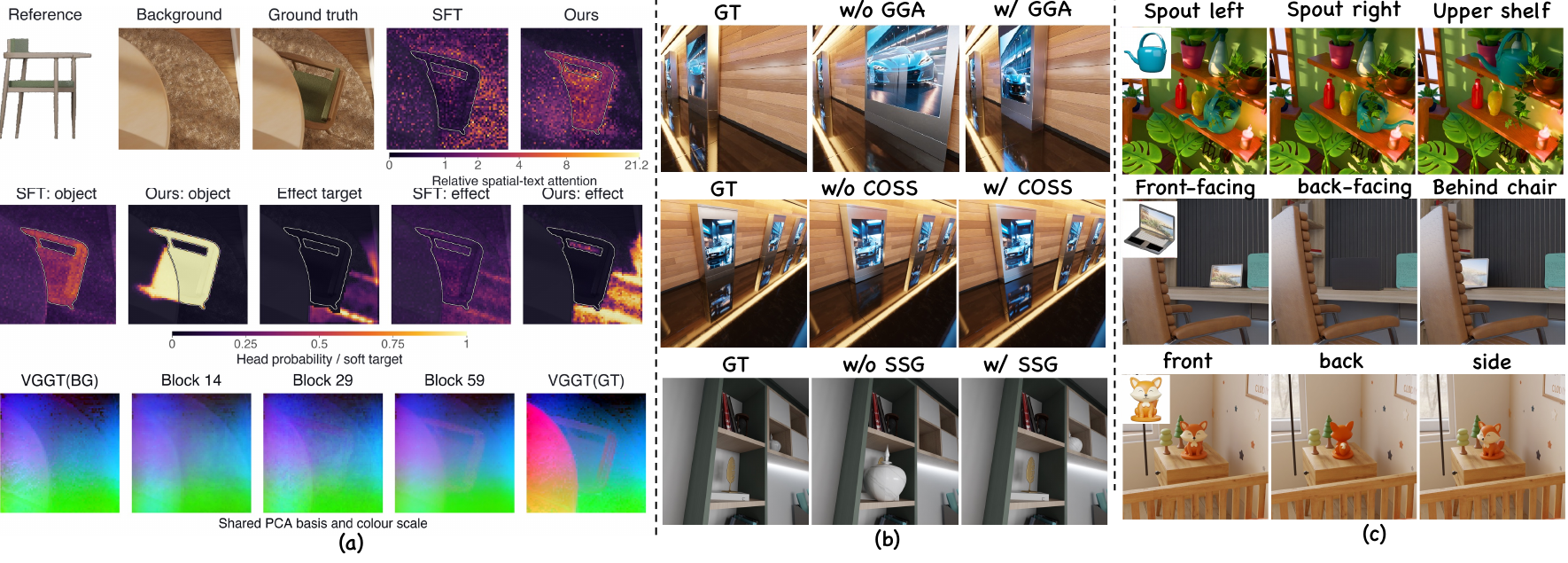}

    \caption{Mechanism probes and qualitative ablations. (a) Spatial-text attention, object/effect predictions, and VGGT feature evolution under identical GT-noised inputs ($\sigma=0.5$). (b) Separate ablations of SSG, GGA, and COSS for placement and scale, perspective, and floor reflection. (c) Text-guided pose and placement control.}
    \label{fig:spatial-attention}
\end{figure}

\begin{table}[htbp]
    \centering
    \caption{Ablations on PhysInsert Bench.}
    \label{tab:ablation}
    \small
    \setlength{\tabcolsep}{4.5pt}
    \renewcommand{\arraystretch}{1.14}
    \resizebox{0.8\textwidth}{!}{%
\begin{tabular}{@{}lcccccccc@{}}
    \toprule
    \textbf{Variant}
    & \shortstack{PSNR$\uparrow$}
    & \shortstack{SSIM$\uparrow$}
    & \shortstack{LPIPS$\downarrow$}
    & \shortstack{FID$\downarrow$}
    & \shortstack{DIoU (\%)$\uparrow$}
    & \shortstack{DINO-Obj.$\uparrow$}
    & \shortstack{AbsRel$\downarrow$}
    & \shortstack{$\delta_{1.25}$ (\%)$\uparrow$} \\
    \midrule
    Init.
    & 19.34 & 0.7840 & 0.2260 & 60.99 & 48.32 & 0.6786 & 0.2207 & 73.35 \\
    PhysW.
    & 22.53 & 0.8529 & 0.1442 & 48.61 & 60.19 & 0.7062 & 0.1372 & 84.18 \\
    w/o GGA
    & 23.83 & 0.8702 & 0.1226 & 43.94 & 61.45 & 0.7206 & 0.1354 & 85.15 \\
    w/o SSG
    & 23.73 & 0.8702 & 0.1203 & 44.97 & \underline{62.51} & 0.7157 & \underline{0.1287} & 86.06 \\
    w/o COSS
    & \underline{23.90} & \underline{0.8722} & \underline{0.1168} & \underline{42.68} & 61.58 & \underline{0.7232} & 0.1315 & \underline{86.40} \\
    Full model
    & \textbf{24.40} & \textbf{0.8773} & \textbf{0.1118} & \textbf{41.96} & \textbf{62.99} & \textbf{0.7276} & \textbf{0.1195} & \textbf{86.56} \\
    \bottomrule
\end{tabular}%
}
\end{table}

\textbf{Quantitative Ablations.} As Table~\ref{tab:ablation}, the full model improves PSNR, DIoU, and LPIPS over PhysW., demonstrating gains beyond dataset fine-tuning. Removing GGA, SSG, or COSS degrades all reported metrics, supporting each component's contribution.

\textbf{Qualitative Ablations.} Figure~\ref{fig:spatial-attention}(b) shows that SSG corrects the white vase's placement, GGA adapts the car poster to the scene's oblique perspective, and COSS strengthens its floor reflection compared with their respective ablations.

\textbf{Spatial Attention.} Under identical text, references, and GT-noised latents, the final model shows more object-focused spatial-text attention than SFT in Figure~\ref{fig:spatial-attention}(a), supporting stronger alignment in this case.

\textbf{Effect Localization.} Under identical GT-noised inputs, the same frozen effect head yields a clearer response in the affected region from final-model features than from SFT features (Figure~\ref{fig:spatial-attention}(a)).

\textbf{Geometry Evolution.} To inspect geometry-stream updates of GGA under the same GT-noised input, we compare auxiliary-head reconstructions from blocks 14, 29, and 59 with background and GT VGGT features in a shared PCA basis, revealing object-region feature changes alongside retained coarse scene structure in this example (Figure~\ref{fig:spatial-attention}(a), bottom).

\textbf{Text-Guided Control.} Figure~\ref{fig:spatial-attention}(c) further shows text-guided changes in laptop orientation and watering-can spout direction, together with placement behind a chair or on an upper shelf.

\chapter{Conclusion}

We presented PhysInsert, combining InsertWorld's counterfactual scene pairs and references across viewpoints and lighting with a VGGT geometry stream and 2D object--scene supervision for joint object and scene adaptation. PhysInsert-Bench and PhysInsert-Bench-Wild evaluate controlled adaptation and real-world generalization. Experiments demonstrate improved image fidelity and placement, supporting joint modeling of objects and their surroundings for physically plausible insertion.

\clearpage
\endgroup

\bibliography{references}
\bibliographystyle{iclr2027_conference}

\clearpage
\appendix
\renewcommand{\theHsection}{appendix.\Alph{section}}
\renewcommand{\theHsubsection}{\theHsection.\arabic{subsection}}
\setcounter{figure}{0}
\setcounter{table}{0}
\setcounter{equation}{0}
\renewcommand{\thefigure}{S\arabic{figure}}
\renewcommand{\thetable}{S\arabic{table}}
\renewcommand{\theequation}{S\arabic{equation}}
\renewcommand{\theHfigure}{supp.\arabic{figure}}
\renewcommand{\theHtable}{supp.\arabic{table}}
\renewcommand{\theHequation}{supp.\arabic{equation}}
\section*{Appendix for PhysInsert: Physically Plausible Object Insertion with Geometric Priors and Bidirectional Object-Scene Interactions}

\chapter{Limitations}

PhysInsert does not explicitly optimize for object identity preservation. Its ability to preserve identity remains limited when the reference and target viewpoints differ substantially. Future work could better exploit complementary information across multiple reference views to preserve distinctive object features under large viewpoint changes.

\section{Geometric Foundation Models}
\label{app:geometric-foundation-models}
VGGT~\citep{VGGT} jointly predicts camera parameters, depth, and 3D scene structure, while Depth Anything 3~\citep{DepthAnything3} recovers consistent geometry across views.
Geometry alone does not determine insertion-induced shadows and reflections, motivating our combination of geometric priors and counterfactual supervision.

\section{More Detailed Quantitative Results}
\label{app:box-guide-subsets}

Table~\ref{tab:box-guide-subsets} reports 180 Optical Adaptation cases, comprising 154 cases with aligned reference--target viewpoints and 26 cross-view Optical challenges, and 220 Pose Adaptation cases comprising 119 Spatial and 101 Mechanical examples. Their single-reference outputs together form the 400-case evaluation in Table~\ref{tab:box-guide}; the three-reference setting covers the same 220 Pose Adaptation cases. +PhysW. baselines receive the same training budget. PhysInsert achieves the best PSNR, LPIPS, and FID on Optical Adaptation, and the best FID, DINO-Obj., and PSNR in both Pose Adaptation reference settings.

\begin{table*}[htbp]
    \centering
    \caption{Box-guided results on 180 Optical Adaptation and 220 Pose Adaptation cases.}
    \label{tab:box-guide-subsets}
    \definecolor{proprietaryblue}{RGB}{224,234,247}
    \definecolor{opensourcepink}{RGB}{253,224,224}
    \setlength{\tabcolsep}{3.5pt}
    \renewcommand{\arraystretch}{1.08}
    \scriptsize
    \resizebox{\textwidth}{!}{%
    \begin{tabular}{lcccccccccc}
        \toprule
        \multirow{3}{*}{\textbf{Method}}
        & \multicolumn{4}{c}{\textbf{Optical Adaptation}}
        & \multicolumn{6}{c}{\textbf{Pose Adaptation}} \\
        \cmidrule(lr){2-5}\cmidrule(lr){6-11}
        & \multicolumn{4}{c}{\textbf{1-view}}
        & \multicolumn{3}{c}{\textbf{1-view}}
        & \multicolumn{3}{c}{\textbf{3-view}} \\
        \cmidrule(lr){2-5}\cmidrule(lr){6-8}\cmidrule(lr){9-11}
        & PSNR$\uparrow$ & SSIM$\uparrow$ & LPIPS$\downarrow$ & FID$\downarrow$
        & FID$\downarrow$ & DINO-Obj.$\uparrow$ & PSNR$\uparrow$
        & FID$\downarrow$ & DINO-Obj.$\uparrow$ & PSNR$\uparrow$ \\
        \midrule
        \rowcolor{proprietaryblue}
        \multicolumn{11}{c}{\textit{\textbf{Proprietary Models}}} \\
        \midrule
        Nano Banana Pro~\citep{NanoBananaPro}
        & 22.92 & 0.8153 & 0.1488 & \underline{56.67}
        & 59.70 & \underline{0.6886} & 22.20
        & \underline{60.85} & \underline{0.6824} & 22.05 \\
        GPT-Image-2~\citep{GPTImage2}
        & 22.01 & 0.8045 & 0.1785 & 66.55
        & 64.31 & 0.6566 & 20.99
        & 66.53 & 0.6404 & 21.06 \\
        \midrule
        \rowcolor{opensourcepink}
        \multicolumn{11}{c}{\textit{\textbf{Open-Source Models}}} \\
        \midrule
        AnyDoor~\citep{Anydoor}
        & 21.08 & 0.8327 & 0.2148 & 82.09
        & 84.99 & 0.5081 & 19.88
        & -- & -- & -- \\
        Insert-Anything~\citep{InsertAnything}
        & 22.79 & 0.8768 & 0.1491 & 68.09
        & 67.83 & 0.6197 & 22.62
        & -- & -- & -- \\
        Insert-Anything + PhysW.~\citep{InsertAnything}
        & \underline{24.44} & 0.8789 & 0.1460 & 57.10
        & \underline{55.32} & 0.6717 & \underline{24.33}
        & 61.72 & 0.6192 & \underline{24.07} \\
        OmniPaint~\citep{Ominipaint}
        & 20.65 & 0.8526 & 0.1337 & 80.98
        & 77.46 & 0.5607 & 20.56
        & -- & -- & -- \\
        OmniPaint + PhysW.~\citep{Ominipaint}
        & 22.73 & 0.8617 & \underline{0.1141} & 68.67
        & 80.24 & 0.5138 & 22.29
        & 86.20 & 0.4570 & 22.06 \\
        FreeCompose~\citep{FreeCompose}
        & 20.88 & 0.7711 & 0.3176 & 83.33
        & 79.59 & 0.4828 & 20.82
        & -- & -- & -- \\
        DIRECT + GT Pose~\citep{Direct}
        & 23.19 & \textbf{0.8975} & 0.1142 & 67.29
        & 61.85 & 0.6422 & 23.10
        & 62.36 & 0.6338 & 23.45 \\
        \midrule
        Qwen-Image-Edit-2511~\citep{QwenImage}
        & 21.44 & 0.8674 & 0.1433 & 83.97
        & 86.51 & 0.4951 & 21.13
        & 146.9 & 0.4406 & 14.93 \\
        \textbf{Ours}
        & \textbf{25.99} & \underline{0.8942} & \textbf{0.09989} & \textbf{49.82}
        & \textbf{45.62} & \textbf{0.7021} & \textbf{25.67}
        & \textbf{44.42} & \textbf{0.7146} & \textbf{25.75} \\
        \bottomrule
    \end{tabular}%
    }
\end{table*}

\FloatBarrier

\section{Evaluation Metric Definitions}
\label{app:evaluation-metrics}

Let $\widehat I_{+}$ and $I_{+}$ denote a generated image and its GT target. Table~\ref{tab:supp-metrics} summarizes the evaluation scope and preferred direction. Except for dataset-level FID, metrics are computed per case and averaged equally over the corresponding evaluation set; DIoU, Mask IoU, and $\delta_{1.25}$ are reported as percentages by multiplying their mean scores by 100. Reference-view settings and evaluation sets follow the respective result tables. All metrics in Table~\ref{tab:text-guide} use one object-reference image per case.

\begin{table}[ht]
    \centering
    \caption{Evaluation metrics and their scope.}
    \label{tab:supp-metrics}
    \small
    \setlength{\tabcolsep}{7pt}
    \begin{tabular}{lll}
        \toprule
        \textbf{Metric} & \textbf{Evaluation scope} & \textbf{Preferred} \\
        \midrule
        PSNR & Full-image pixel fidelity & Higher \\
        SSIM & Full-image structural similarity & Higher \\
        LPIPS & Full-image perceptual distance & Lower \\
        FID & Full-image feature distributions & Lower \\
        DIoU (\%) & Mask-box overlap and center distance & Higher \\
        Mask IoU (\%) & Visible foreground overlap & Higher \\
        DINO-Obj. & Object-crop identity similarity & Higher \\
        Phy-VLM & Physical plausibility with case-specific focus & Higher \\
        AbsRel & Valid object-region depth error & Lower \\
        $\delta_{1.25}$ (\%) & Valid object-region depth agreement & Higher \\
        \bottomrule
    \end{tabular}
\end{table}

\subsection{Full-Image Fidelity}

\paragraph{PSNR.}
Peak signal-to-noise ratio measures pixel fidelity over the complete RGB image. For images of size $H\times W$ and pixel dynamic range $R$, it is
\begin{equation}
\begin{aligned}
    \operatorname{MSE} &= \frac{1}{3HW}\sum_{u=1}^{H}\sum_{v=1}^{W}\sum_{c=1}^{3}
    \left(\widehat I_{+}(u,v,c)-I_{+}(u,v,c)\right)^2, \\
    \operatorname{PSNR} &= 10\log_{10}\!\left(\frac{R^2}{\operatorname{MSE}}\right).
\end{aligned}
\label{eq:supp-psnr}
\end{equation}
Our implementation uses RGB values in $[0,255]$ and $R=255$.

\paragraph{SSIM.}
Structural similarity~\citep{SSIM} compares local luminance, contrast, and structure. For corresponding local windows $x$ and $y$ from the generated and GT images,
\begin{equation}
    \operatorname{SSIM}(x,y)=
    \frac{(2\mu_x\mu_y+C_1)(2\sigma_{xy}+C_2)}
    {(\mu_x^2+\mu_y^2+C_1)(\sigma_x^2+\sigma_y^2+C_2)},
\label{eq:supp-ssim}
\end{equation}
where $\mu$, $\sigma^2$, and $\sigma_{xy}$ are local means, variances, and covariance; $C_1$ and $C_2$ stabilize the ratios. We use the scikit-image implementation with RGB channel averaging and a dynamic range of 255, averaging local scores over the image.

\paragraph{LPIPS.}
Learned Perceptual Image Patch Similarity~\citep{LPIPS} measures perceptual distance using deep features. Let $\widetilde\phi_\ell(I)$ denote channel-normalized features at layer $\ell$ and $w_\ell$ the learned channel weights. Then
\begin{equation}
    \operatorname{LPIPS}(\widehat I_{+},I_{+})=
    \sum_{\ell}\frac{1}{H_\ell W_\ell}\sum_{u,v}
    \left\|w_\ell\odot\left(
    \widetilde\phi_\ell(\widehat I_{+})_{uv}
    -\widetilde\phi_\ell(I_{+})_{uv}\right)\right\|_2^2.
\label{eq:supp-lpips}
\end{equation}
We use the AlexNet-based LPIPS implementation on complete images, with RGB inputs mapped to $[-1,1]$.

\paragraph{FID.}
Fr\'echet Inception Distance~\citep{FID} compares the feature distributions of generated and GT full-image sets. With feature means $\mu_g,\mu_t$ and sample covariances $\Sigma_g,\Sigma_t$,
\begin{equation}
    \operatorname{FID}=
    \|\mu_g-\mu_t\|_2^2+
    \operatorname{Tr}\!\left(
    \Sigma_g+\Sigma_t-2(\Sigma_g^{1/2}\Sigma_t\Sigma_g^{1/2})^{1/2}
    \right).
\label{eq:supp-fid}
\end{equation}
FID is computed once over all generated and GT images in each reported evaluation set, including by concatenating features across evaluation shards; it is not an average of per-image scores. The implementation evaluates the covariance cross term through an equivalent low-rank singular-value decomposition.

\subsection{Text-Guided Placement Accuracy}

\label{app:placement-metrics}

\paragraph{Foreground extraction for Table~\ref{tab:text-guide}.}
We apply one fixed evaluator to all methods on the same $N=400$ original text-guided cases. Qwen3.5-Plus receives, in order, the input background, the object-reference collage, and the generated image. Using the editing instruction and a component vocabulary fixed per case across methods, it localizes all visible newly inserted target instances and parts, excluding pre-existing background objects, cast shadows, and reflections. Reference panels depict views of the same target assembly rather than additional copies. The grounder returns component boxes in $xyxy$ coordinates normalized to $[0,1000]$, with temperature 0 and a requested maximum of 8192 output tokens. Neither GT RGB images nor GT masks or boxes are provided to localization or segmentation.

RGB images are resized bilinearly to a shared canvas preserving the background aspect ratio, with longest side at most 1024 pixels. The reference collage preserves its own aspect ratio with the same size cap. VLM inputs are encoded as RGB JPEGs with quality 95 and no chroma subsampling. SAM 3~\citep{SAM3} (\texttt{facebook/sam3}) generates candidates for each component phrase, using score and mask-probability thresholds of 0.5 and discarding masks smaller than 16 pixels. For each VLM box, we select every same-component candidate whose tight-box center lies inside that box and at least half of whose mask pixels lie inside it. If no candidate meets both conditions, we select the existing same-component candidate with highest box DIoU to the VLM box, retaining negative match scores. If that component has no candidates, we retry its category head noun (preserving ``bean bag'' as a phrase); if still unmatched, SAM3's box-prompt tracker supplies a single mask with logit threshold 0. Selected masks are united into $\widehat M_n$. Model weights remain in float32 with bfloat16 autocasting and TF32 enabled. There is no GT-based mask selection, per-method tuning, or manual mask correction. Automatic localization and segmentation can still omit parts or select background regions.

\paragraph{Mask IoU.}
Let $M_n$ be the GT visible-object mask, resized to the same canvas by nearest-neighbor interpolation and binarized at values greater than 127. We report
\begin{equation}
    \operatorname{MaskIoU}(\%)=
    \frac{100}{N}\sum_{n=1}^{N}
    \frac{|\widehat M_n\cap M_n|}{|\widehat M_n\cup M_n|+10^{-6}}.
\label{eq:supp-mask-iou}
\end{equation}
This measures foreground overlap rather than enclosing-box overlap. The GT visible mask is used only at this scoring stage.

\paragraph{DIoU.}
Let $\widehat B_n$ and $B_n$ be tight, axis-aligned boxes of $\widehat M_n$ and $M_n$, using half-open pixel coordinates. Let $\widehat{\mathbf c}_n$ and $\mathbf c_n$ denote their centers, and $d_n$ the diagonal length of the smallest box enclosing both. For a nonempty prediction, the distance-IoU score~\citep{DIoU} is
\begin{equation}
\begin{aligned}
    u_n &= \frac{|\widehat B_n\cap B_n|}{|\widehat B_n\cup B_n|+10^{-8}}, \\
    s_n &= u_n-\frac{\|\widehat{\mathbf c}_n-\mathbf c_n\|_2^2}{d_n^2+10^{-8}},
    \qquad \operatorname{DIoU}(\%)=\frac{100}{N}\sum_{n=1}^{N}s_n.
\end{aligned}
\label{eq:supp-bbox-diou}
\end{equation}
An empty predicted mask receives zero for both metrics and remains in the 400-case average. Negative DIoU values are retained: a displaced box can incur a center-distance penalty larger than its overlap. Higher values are better for both metrics. The JoyAI row uses the existing \texttt{text400} outputs from JoyAI-Image-Edit-Plus-Diffusers, as in the depth evaluation below.

Tables~\ref{tab:text-guide} and~\ref{tab:ablation} use the same percentage-scaled DIoU definition in Eq.~\eqref{eq:supp-bbox-diou}.

\subsection{Object Identity Preservation}

\paragraph{DINO-Obj.}
DINO-Obj. measures identity preservation by comparing object crops from the generated and GT images. We use the evaluator's GroundingDINO-selected~\citep{liu2024grounding} prediction box and the GT object box, expand both with the same crop factor, and crop the respective RGB images while retaining context inside each box. The crop evaluator uses detection confidence and the fraction of changed pixels relative to $I_{-}$ to select its prediction box; the SAM3-based placement metrics above are evaluated separately and do not change these DINO-Obj. crops. The crop-expansion factor is 1.1.
Let $C_{g,n}$ and $C_{t,n}$ denote the generated-image and GT object crops for case $n$, and let $f$ extract the CLS feature from frozen DINOv2~\citep{DINOv2} with registers~\citep{ViTRegisters} (\texttt{facebook/dinov2-with-registers-base}). We define the per-case score $s_n^{\mathrm{DINO}}$ and the dataset average as
\begin{equation}
\begin{aligned}
    s_n^{\mathrm{DINO}} &=
    \frac{f(C_{g,n})^{\mathsf T}f(C_{t,n})}
    {\|f(C_{g,n})\|_2\,\|f(C_{t,n})\|_2}, \\
    \operatorname{DINO\text{-}Obj.} &= \frac{1}{N}\sum_{n=1}^{N}s_n^{\mathrm{DINO}}.
\end{aligned}
\label{eq:supp-dino-obj}
\end{equation}
Here $n\in\{1,\ldots,N\}$ indexes all evaluated cases, including missed detections, for which $s_n^{\mathrm{DINO}}=0$. Features are $\ell_2$-normalized before computing the similarity. The comparison uses the GT target-object crop rather than the reference-view collage, and pixels outside the object silhouette but inside the crop are retained.

\subsection{Object-Region Depth Consistency}
\label{app:depth-metrics}

Following the depth-metric definitions and paired-input procedure described in PhyEdit~\citep{PhyEdit}, we use frozen Depth Anything 3 (DA3NESTED-GIANT-LARGE)~\citep{DepthAnything3}. Let $I_{-}$ be the background, $I_{+}$ the GT target, and $\widehat I_{+}$ the generated image. We estimate depths in two separate forward passes:
\begin{equation}
\begin{aligned}
    (D_{-}^{g},D_g) &= \operatorname{DA3}([I_{-},I_{+}]), \\
    (D_{-}^{p},D_p) &= \operatorname{DA3}([I_{-},\widehat I_{+}]).
\end{aligned}
\label{eq:depth-paired-inputs}
\end{equation}
Both branches use the same model weights, preprocessing, and per-case random seed, and we compare the second depth output from each pair. The generated-image branch receives only the background and generated image; the GT image is used to obtain the reference depth for scoring.

All RGB inputs are resized to a common canvas preserving the target aspect ratio, with a longest edge of 768 pixels; the GT object mask uses nearest-neighbor resizing. When needed, depth outputs are resized to this canvas by bilinear interpolation and validity masks by nearest-neighbor interpolation. For case $n$, let $d_{n,i}=D_{g,n}(i)$ and $\hat d_{n,i}=D_{p,n}(i)$. The evaluation region $\Omega_n$ is the GT object mask intersected with the non-sky validity masks from both outputs, retaining only pixels with finite $d_{n,i},\hat d_{n,i}>10^{-6}$. Corresponding pixel coordinates are compared directly, without additional depth-scale or offset fitting.

\begin{samepage}
\paragraph{AbsRel and $\delta_{1.25}$.}
For each case, absolute relative error measures the mean absolute depth difference relative to the estimated target depth, and threshold accuracy measures the fraction of valid object pixels passing the ratio threshold:
\begin{equation}
\begin{aligned}
    \operatorname{AbsRel}_n &= \frac{1}{|\Omega_n|}\sum_{i\in\Omega_n}\frac{|\hat d_{n,i}-d_{n,i}|}{d_{n,i}}, \\
    \delta_{1.25,n} &= \frac{1}{|\Omega_n|}\sum_{i\in\Omega_n}
    \mathbf{1}\!\left[\max\!\left(\frac{\hat d_{n,i}}{d_{n,i}},\frac{d_{n,i}}{\hat d_{n,i}}\right)<1.25\right].
\end{aligned}
\label{eq:depth-metrics}
\end{equation}
\end{samepage}
The threshold is strict: pixels with a depth ratio equal to 1.25 do not pass. Lower AbsRel and higher $\delta_{1.25}$ indicate better agreement.

Within each setting, every method is evaluated on the same $N=400$ cases: the text-guided outputs for Table~\ref{tab:text-guide}, and the single-view box-guided outputs for Table~\ref{tab:box-guide} (180 Optical Adaptation and 220 Pose Adaptation cases). All cases have valid object pixels. We report equally weighted per-image averages:
\begin{equation}
    \operatorname{AbsRel}=\frac{1}{N}\sum_{n=1}^{N}\operatorname{AbsRel}_n,
    \qquad
    \delta_{1.25}(\%)=\frac{100}{N}\sum_{n=1}^{N}\delta_{1.25,n}.
\label{eq:depth-aggregation}
\end{equation}
The JoyAI row uses the existing \texttt{text400} outputs from JoyAI-Image-Edit-Plus-Diffusers for depth evaluation. Both target and generated depths are DA3 estimates. These scores therefore measure estimated depth consistency and remain sensitive to depth-estimator errors and paired-image context.

\subsection{VLM-Based Physical Plausibility}
\label{app:phy-vlm}

We adapt the VLM-based physical plausibility assessment in PhyEdit~\citep{PhyEdit} to PhysInsert-Bench, using GPT-5.5 to evaluate the outputs of all compared methods on the same 400 cases. Each evaluation prompt combines a fixed template with the case's annotated physical effects and evaluation-focus tags from Appendix~\ref{app:benchmark-construction}. The tags specify which physical effects warrant particular attention, such as illumination adaptation, cast shadows, reflections, or occlusion relationships, as applicable to the case. The same template and case-specific tag information are used across methods. For each method and case, we obtain three valid numerical ratings, retrying any evaluation that returns \texttt{null} until three valid ratings are available. No case is excluded because of a \texttt{null} response. We first average the three valid ratings within each case, then average these case-level scores equally over all 400 cases. We report the resulting Phy-VLM mean with the half-width of its 95\% confidence interval; higher values indicate greater physical plausibility as judged by the VLM.

\paragraph{Scoring prompt.}
The following is an English translation of the scoring prompt used for Phy-VLM in Table~\ref{tab:text-guide}. The placeholders contain the case-specific evaluation focus and the original insertion instruction.

\begin{tcolorbox}[
  enhanced,breakable,
  colback=black!2,colframe=black!28,
  colbacktitle=black!7,coltitle=black,
  boxrule=0.4pt,arc=1.3mm,
  left=9pt,right=9pt,top=7pt,bottom=7pt,
  before skip=10pt,after skip=10pt,
  fonttitle=\small\bfseries,
  fontupper=\fontsize{9}{11}\selectfont,
  pad at break*=2mm,
  title={Phy-VLM scoring prompt (English translation)}
]
\setlength{\parindent}{0pt}
\setlength{\parskip}{5pt}
Evaluate the physical plausibility of the edited result relative to the ground truth (GT). Do not evaluate aesthetics, identity preservation, or pixel similarity, and do not speculate about the generation method.

Image 1: edited result; Image 2: GT; Image 3: original background; Image 4: reference object (multiple views still depict the same object). The background and reference object are provided only to aid understanding.

First compare against the GT and separately verify the physical relationships in the GT and the result. Do not assume that a relationship present in the GT is also present in the result.

Evaluation focus: \texttt{\{\{CASE\_FOCUS\_TEXT\}\}}

Original insertion instruction (in this instruction, Image 1 refers to the background and Image 2 to the reference object, rather than the scoring-image order above): \texttt{\{\{ORIGINAL\_INSTRUCTION\}\}}

Incorrect object size alone does not constitute physical implausibility: do not deduct points merely because the object is larger or smaller than in the GT, relative to neighboring objects, or than specified by the instruction. This rule takes precedence over size requirements in the case annotations and the instruction.

Evaluate only gravitational relationships (support, contact, and hanging), optical relationships (illumination, environmental color, shadows, reflections, and transmission), and geometric relationships (perspective, orientation, occlusion, and interpenetration). Do not separately penalize object size or changes in shadow or reflection extent that naturally result from it; do not infer errors in these relationships solely from a size difference.

Compare with the GT according to the evaluation focus:\par
\textbf{5}: Key relationships are confirmed to be consistent or physically equivalent.\par
\textbf{4}: Key relationships are retained, with only minor local deviations.\par
\textbf{3}: The main relationships hold, but important effects clearly deviate.\par
\textbf{2}: Key relationships deviate substantially, with only partial consistency.\par
\textbf{1}: Core relationships are missing, reversed, or fundamentally broken.\par
Large discrepancies from the GT in key physical relationships must receive a low score; appearance or size differences alone do not count.

A clearly visible, severe gravitational, optical, or geometric error outside the stated focus caps the score at 2. Assign 1 for an entirely unsupported floating object, a disconnected primary load-bearing contact, obvious interpenetration, or impossible occlusion. Minor issues outside the focus do not reduce the score. Proximity or overlapping contours do not establish load-bearing support; any deduction must be supported by visible evidence.

Return \texttt{null} when key evidence is unclear or the GT contains an obvious error that affects the judgment; do not automatically treat such cases as good or bad.

Return only JSON: \texttt{\{"score":} an integer from 1 to 5 or \texttt{null, "reason": "}One sentence identifying the key physical difference between the result and the GT, or stating the evidence for consistency or the need for review.\texttt{"\}}.
\end{tcolorbox}

\section{Method Details and Notation}
\label{app:method-notation}
\label{app:method-details}

We use $Z$ for latents, $U_\ell$, $T_\ell$, and $G_\ell$ for image-, text-, and geometry-stream states, and $F$ for VGGT features. $N_r$ is the number of reference images, $L$ the number of three-stream DiT blocks, and $\ell$ the layer/state index; $\mathcal{B}_\ell$ denotes the corresponding update block. The sets $\mathcal{S}_{\mathrm{geo}}$, $\mathcal{S}_{\mathrm{cf}}$, and $\mathcal{S}_{\mathrm{ground}}$ select the layers used for geometry readout, counterfactual spatial prediction, and grounding supervision, respectively; $|\mathcal{S}|$ denotes a set's cardinality.

\subsection{Latent Encoding and Flow Matching}
\label{app:flow-details}

A frozen VAE encodes the target image $I_{+}$, background $I_{-}$, and each reference $I_r^i$ into $Z_0$, $Z_{-}$, and $Z_r^i$, respectively. The reference latents are concatenated as $Z_r=[Z_r^1;\ldots;Z_r^{N_r}]$. A frozen multimodal encoder produces the initial text tokens $T_0$ from instruction $p$ and the image context. The image stream is initialized as $U_0=[Z_t;Z_{-};Z_r]$. During training, we sample noise $\epsilon$, form the noisy target latent, and optimize
\begin{equation}
\begin{aligned}
    Z_t&=(1-\sigma_t)Z_0+\sigma_t\epsilon, \\
    \mathcal{L}_{\mathrm{flow}}
    &=\mathbb{E}_{t,\epsilon}\!\left[
    w(t)\left\|v_\theta(Z_t;I_{-},\mathcal{I}_r,p)-(\epsilon-Z_0)\right\|_2^2
    \right].
\end{aligned}
\label{eq:supp-flow-matching}
\end{equation}

Here $t$ is the sampled diffusion timestep, $\sigma_t$ the noise-mixing coefficient, $w(t)$ the timestep-dependent loss weight, and $v_\theta$ the predicted velocity field. The coefficients $\sigma_t$ and $w(t)$ follow the backbone's flow-matching scheduler~\citep{FlowMatching}. We adapt the image and text branches with LoRA and train the added geometry branch, while keeping the VAE and multimodal encoder frozen.

\subsection{Geometry Features and Alignment}
\label{app:geometry-details}

The frozen VGGT~\citep{VGGT} extracts background features $F_{-}=f_{\mathrm{VGGT}}^{(\ell_v)}(I_{-})$ at layer $\ell_v$. The MLP $A_\phi$ maps these $d_G$-dimensional features to the $d_D$-dimensional DiT~\citep{DiT} hidden space, and the learned geometry-modality embedding $e_G$ gives $G_0=A_\phi(F_{-})+e_G$. Each stream computes its own queries, keys, and values, which are concatenated for joint attention, and updates its states through stream-specific modulation, output projections, and feed-forward paths. Motivated by similarities between depth and RGB features, the geometry branch reuses the image branch's linear weights as a frozen base and applies independent low-rank adapters. The geometry states evolve alongside the image and text states through the $L$ three-stream blocks and their own residual updates.

The same frozen VGGT extracts target features $F_{+}=\operatorname{sg}(f_{\mathrm{VGGT}}^{(\ell_v)}(I_{+}))$, where $\operatorname{sg}$ stops gradient propagation. A shared readout head $P_\psi$ predicts these features from geometry states, with $\widehat F_{+}^{(\ell)}=P_\psi(G_\ell)$. The distance used in the main-text geometry objective is
\begin{equation}
\begin{aligned}
    \mathcal{D}_{\mathrm{geo}}(A,B)
    &=\operatorname{SL1}_{\beta_{\mathrm{geo}}}\!\left(A,\operatorname{LN}(B)\right) \\
    &\quad+\lambda_{\mathrm{cos}}\operatorname{CosDist}\!\left(A,\operatorname{LN}(B)\right), \\
    \mathcal{L}_{\mathrm{geo}}
    &=\sum_{\ell\in\mathcal{S}_{\mathrm{geo}}}\omega_\ell
    \mathcal{D}_{\mathrm{geo}}\!\left(\widehat F_{+}^{(\ell)},F_{+}\right).
\end{aligned}
\label{eq:supp-geometry-distance}
\end{equation}

The projector and readout have trainable parameters $\phi$ and $\psi$. Here $\operatorname{LN}$ denotes layer normalization; $\beta_{\mathrm{geo}}>0$ controls the transition between the quadratic and linear regions of Smooth L1, and $\lambda_{\mathrm{cos}}$ weights the cosine-distance term. The layer weights satisfy $\omega_\ell\geq0$ and $\sum_{\ell\in\mathcal{S}_{\mathrm{geo}}}\omega_\ell=1$. Target features serve only as training targets and are never injected into the generator.

\subsection{Counterfactual Targets and Spatial Prediction}
\label{app:counterfactual-details}

We downsample the ground-truth object mask to the target-token grid to obtain the soft object occupancy map $M_o$. From aligned with/without images, the difference map $D$ and external-effect map $M_e$ are
\begin{align}
    D &= \operatorname{AvgPool}_{\kappa\times\kappa}\!\left(
    \operatorname{Mean}_{\mathrm{RGB}}
    \left|\operatorname{Down}(I_{+})-\operatorname{Down}(I_{-})\right|
    \right), \\
    M_e &= \operatorname{clip}\!\left(
    \frac{D-\delta_{\mathrm{low}}}{\delta_{\mathrm{high}}-\delta_{\mathrm{low}}},0,1\right)
    \left[1-\operatorname{Dilate}_{\rho_d}\!\left(\mathbf{1}[M_o>\delta_{\mathrm{obj}}]\right)\right].
\label{eq:supp-counterfactual-targets}
\end{align}

All three maps are defined on the target-token grid. $\operatorname{Down}$ downsamples an image to this grid, $\operatorname{Mean}_{\mathrm{RGB}}$ averages the absolute differences over RGB channels, and $\kappa$ is the side length of the average-pooling kernel. The thresholds $\delta_{\mathrm{low}}<\delta_{\mathrm{high}}$ map a difference magnitude to a soft effect strength: values below the lower threshold map to zero, values above the upper threshold map to one, and intermediate values are linearly rescaled. $\delta_{\mathrm{obj}}$ binarizes object occupancy, and $\operatorname{Dilate}_{\rho_d}$ expands that region by radius $\rho_d$ in grid coordinates to exclude the object and its boundary neighborhood. $\mathbf{1}[\cdot]$ is an indicator and $\operatorname{clip}(\cdot,0,1)$ clamps its argument to the unit interval. Thus, $M_e$ is a proxy for visible changes outside the object, including cast shadows and reflections, rather than a semantic decomposition of individual optical effects.

An object--effect spatial head is shared across target-image tokens at layers in $\mathcal{S}_{\mathrm{cf}}$. After averaging its logits across layers, $\widehat M_c$ denotes the predicted probability map for channel $c\in\{o,e\}$, corresponding to object occupancy and external effects. Each channel combines balanced binary cross-entropy and Dice losses:
\begin{equation}
    \mathcal{L}_c=
    \alpha_{\mathrm{bce}}\mathcal{L}_{\mathrm{bce}}(\widehat M_c,M_c)
    +\alpha_{\mathrm{dice}}\mathcal{L}_{\mathrm{dice}}(\widehat M_c,M_c).
\label{eq:supp-counterfactual-channel-loss}
\end{equation}
The nonnegative weights $\alpha_{\mathrm{bce}}$ and $\alpha_{\mathrm{dice}}$ sum to one. The main-text counterfactual objective combines the two channels as $\mathcal{L}_{\mathrm{cf}}=\alpha_o\mathcal{L}_o+\alpha_e\mathcal{L}_e$, with $\alpha_o,\alpha_e\geq0$ and $\alpha_o+\alpha_e=1$.

\subsection{Spatial-Text Attention and Mask Alignment}
\label{app:grounding-details}

The generated caption contains a dedicated spatial-description field, shown in Figure~\ref{fig:dataset-construction}. We use that field to identify the corresponding text-token subset $\mathcal{T}_s$ in the instruction. Indices $i$ and $i'$ range over target-image token positions, whereas $j$ indexes $\mathcal{T}_s$. At layer $\ell$ and head $h$, $q_i^{(\ell,h)}$ is the target-image query and $k_j^{(\ell,h)}$ the spatial-text key; each has dimension $d_h$, and there are $N_h$ heads. The spatial relevance score aggregates over the phrase tokens using log-mean-exp and then averages over heads:
\begin{equation}
    r_i^{(\ell)}
    =\frac{1}{N_h}\sum_{h=1}^{N_h}
    \log\!\left[
    \frac{1}{|\mathcal{T}_s|}
    \sum_{j\in\mathcal{T}_s}
    \exp\!\left(\frac{q_i^{(\ell,h)}\cdot k_j^{(\ell,h)}}{\sqrt{d_h}}\right)
    \right].
\label{eq:supp-spatial-relevance}
\end{equation}

The positive temperature $\tau_a$ controls the concentration of the attention distribution, whose softmax is taken over target-image positions only. We normalize the soft object mask on the same grid and compute
\begin{align}
    P_r^{(\ell)}(i)&=\operatorname{softmax}_i(r_i^{(\ell)}/\tau_a),
    \qquad
    P_o(i)=\frac{M_o(i)}{\sum_{i'}M_o(i')}, \\
    \mathcal{L}_{\mathrm{ground}}
    &=\frac{1}{|\mathcal{S}_{\mathrm{ground}}|}
    \sum_{\ell\in\mathcal{S}_{\mathrm{ground}}}
    D_{\mathrm{KL}}\!\left(P_o\,\middle\|\,P_r^{(\ell)}\right).
\label{eq:supp-grounding-alignment}
\end{align}
Here $D_{\mathrm{KL}}$ denotes Kullback--Leibler divergence. This objective supervises the spatial distribution of image--text relevance. The object mask supplies training supervision and is not required at inference.

\paragraph{Overall loss weights.}
The overall coefficients $\lambda_{\mathrm{geo}}$, $\lambda_{\mathrm{cf}}$, and $\lambda_{\mathrm{ground}}$ weight the geometry, counterfactual, and grounding objectives relative to $\mathcal{L}_{\mathrm{flow}}$. They are nonnegative and are not constrained to sum to one, unlike the normalized layer and channel weights within the auxiliary objectives. The 2D-OSS loss groups the weighted counterfactual and grounding terms as $\mathcal{L}_{\mathrm{2D}}=\lambda_{\mathrm{cf}}\mathcal{L}_{\mathrm{cf}}+\lambda_{\mathrm{ground}}\mathcal{L}_{\mathrm{ground}}$, preserving these coefficients.

\clearpage
\section{Training and Inference Hyperparameters}
\label{app:hyperparameters}

We use AdamW~\citep{AdamW}, which applies decoupled weight decay to the adaptive-moment updates of Adam~\citep{Adam}. Tables~\ref{tab:training-inference-hyperparameters} and~\ref{tab:loss-hyperparameters} report the training, inference, and auxiliary-supervision settings of the base PhysInsert model. The base PhysInsert model is trained in a single stage for approximately 16,000 optimizer updates using the joint objective in Eq.~(\ref{eq:overall-objective}). The box-guided model is obtained by fine-tuning the text-guided model for an additional 4,000 steps. The +PhysW. baselines use the same training budget in each setting, including the additional adaptation for box-guided evaluation.

\begin{table}[htbp]
    \centering
    \caption{Base-model training and inference hyperparameters. The additional box-guided training budget is listed separately.}
    \label{tab:training-inference-hyperparameters}
    \small
    \renewcommand{\arraystretch}{1.12}
    \begin{tabular}{@{}p{0.49\linewidth}p{0.46\linewidth}@{}}
        \toprule
        Setting & Value \\
        \midrule
        \multicolumn{2}{@{}l}{\textit{Base-model training}} \\
        Optimizer & AdamW \\
        Base learning rate & $1\times10^{-4}$ \\
        Weight decay & $0.01$ \\
        Learning-rate scheduler & PyTorch \texttt{ConstantLR} (defaults) \\
        Effective batch size & $8$ \\
        Batch size per GPU / accumulation steps & $1$ / $1$ \\
        Training hardware & $8\times$ NVIDIA L20X \\
        Mixed precision & BF16 \\
        LoRA rank (image, text, and geometry) & $32$ \\
        Maximum training-image pixels & $1{,}048{,}576$ ($1024^2$) \\
        Training schedule & Single-stage joint optimization \\
        Total base-model updates & Approximately $16{,}000$ \\
        \midrule
        \multicolumn{2}{@{}l}{\textit{Additional box-guided adaptation}} \\
        Additional optimizer updates & $4{,}000$ \\
        \midrule
        \multicolumn{2}{@{}l}{\textit{Inference}} \\
        Classifier-free guidance (CFG) scale & $4.0$ \\
        Denoising steps & $40$ \\
        Sampler & Euler flow-matching solver \\
        Noise schedule & Qwen-Image dynamic shift \\
        Denoising strength & $1.0$ \\
        Negative prompt & Empty \\
        Maximum output pixels & $1{,}048{,}576$ ($1024^2$) \\
        \bottomrule
    \end{tabular}
\end{table}
\FloatBarrier

We jointly optimize flow matching, geometry alignment, COSS, and SSG in a single training stage using Eq.~(\ref{eq:overall-objective}). The auxiliary-loss weights are listed below; all layer indices are zero-based.

\begin{table}[htbp]
    \centering
    \caption{Auxiliary-loss weights and module hyperparameters for single-stage training.}
    \label{tab:loss-hyperparameters}
    \small
    \setlength{\tabcolsep}{4pt}
    \renewcommand{\arraystretch}{1.12}
    \begin{tabular}{@{}p{0.62\linewidth}c@{}}
        \toprule
        Parameter & Value \\
        \midrule
        Flow Matching Loss coefficient & $1$ \\
        Geometry Alignment Loss, $\lambda_{\mathrm{geo}}$ & $0.02$ \\
        Counterfactual Supervision Loss, $\lambda_{\mathrm{cf}}$ & $0.01$ \\
        Object Grounding Loss, $\lambda_{\mathrm{ground}}$ & $0.005$ \\
        \midrule
        \multicolumn{2}{@{}l}{\textit{Module settings}} \\
        VGGT feature layer, $\ell_v$ & \multicolumn{1}{c}{$23$} \\
        Feature / DiT dimensions, $d_G$ / $d_D$ & \multicolumn{1}{c}{$2048$ / $3072$} \\
        Projector / readout MLP hidden width & \multicolumn{1}{c}{$256$ / $256$} \\
        GGA readout layers, $\mathcal{S}_{\mathrm{geo}}$ & \multicolumn{1}{c}{$\{14,29,44,59\}$} \\
        GGA normalized layer weights, $\omega_\ell$ & \multicolumn{1}{c}{$(1,2,4,10)/17$} \\
        Smooth L1 transition, $\beta_{\mathrm{geo}}$ & \multicolumn{1}{c}{$0.1$} \\
        Cosine-distance weight, $\lambda_{\mathrm{cos}}$ & \multicolumn{1}{c}{$0.25$} \\
        COSS / SSG layers, $\mathcal{S}_{\mathrm{cf}}=\mathcal{S}_{\mathrm{ground}}$ & \multicolumn{1}{c}{$\{35,47,56\}$} \\
        Object / effect weights, $\alpha_o$ / $\alpha_e$ & \multicolumn{1}{c}{$0.65$ / $0.35$} \\
        BCE / Dice weights, $\alpha_{\mathrm{bce}}$ / $\alpha_{\mathrm{dice}}$ & \multicolumn{1}{c}{$0.5$ / $0.5$} \\
        Balanced-BCE positive-weight cap & \multicolumn{1}{c}{$20$} \\
        Effect thresholds, $\delta_{\mathrm{low}}$ / $\delta_{\mathrm{high}}$ & \multicolumn{1}{c}{$0.02$ / $0.10$} \\
        Object threshold, $\delta_{\mathrm{obj}}$ & \multicolumn{1}{c}{$0.01$} \\
        Smoothing-kernel side length, $\kappa$ & \multicolumn{1}{c}{$3$} \\
        Exclusion dilation radius, $\rho_d$ & \multicolumn{1}{c}{$1$ token-grid cell} \\
        Grounding temperature, $\tau_a$ & \multicolumn{1}{c}{$1.0$} \\
        \bottomrule
    \end{tabular}
\end{table}
\FloatBarrier

Effect thresholds operate on RGB values normalized to $[0,1]$, while smoothing and dilation operate on the target-token grid.

\clearpage
\section{Inference Efficiency}
\label{app:inference-efficiency}

\begin{table}[htbp]
    \centering
    \caption{Inference efficiency on one NVIDIA L20X GPU at $1024\times1024$,
    batch size 1, 40 denoising steps, and CFG 4.0. Parameter counts are obtained
    from the loaded inference modules. Latency is the mean and sample standard
    deviation over five cases after two warmup calls; memory is the maximum
    PyTorch-allocated GPU memory over the timed calls. B denotes $10^9$ parameters.}
    \label{tab:inference-efficiency}
    \small
    \setlength{\tabcolsep}{7pt}
    \renewcommand{\arraystretch}{1.15}
    \begin{tabular}{@{}lrrrr@{}}
        \toprule
        Method & \shortstack{Denoiser\\params (B)} & \shortstack{Pipeline\\params (B)}
        & \shortstack{Time/image\\(s)} & \shortstack{Peak memory\\(GiB)} \\
        \midrule
        Qwen-Image-Edit-2511 & 20.430 & 28.849 & $79.40\pm0.06$ & 55.92 \\
        Ours (MLP) & 20.815 & 30.143 & $120.13\pm0.26$ & 60.80 \\
        \bottomrule
    \end{tabular}
\end{table}

\paragraph{Parameter accounting.}
We count parameters in the loaded MLP-based PhysInsert model.
The $2048\rightarrow256\rightarrow3072$ geometry MLP contains 1,314,048
parameters; including input LayerNorm and the geometry-modality embedding gives
1,321,216 parameters. Denoiser counts include the DiT, its LoRA parameters,
the geometry projector, and third-stream adapters. Pipeline counts additionally
include the multimodal encoder, VAE, and frozen VGGT aggregator. Shared weights
are counted once, and training-only feature readout and spatial heads are excluded.
This gives approximately 20.815B denoiser and 30.143B pipeline
parameters, a 4.48\% pipeline increase over the backbone.

\paragraph{Measurement protocol.}
Both models use the same NVIDIA L20X GPU, BF16 weights, PyTorch 2.12.1,
CUDA 12.6, and no model offloading, with identical background, reference,
and box-guidance images, prompts, and seeds.
CUDA synchronization brackets each complete pipeline call, including encoding,
denoising, and decoding; file I/O and model loading are excluded.
PhysInsert measurements include online VGGT extraction and unfused LoRA execution.

\FloatBarrier

\clearpage
\section{More Qualitative Results}
\label{app:more-qualitative-results}

\begin{figure}[!ht]
    \centering
    \begingroup
    \setlength{\tabcolsep}{0pt}
    \scriptsize
    \begin{tabular}{@{}*{10}{>{\centering\arraybackslash}p{0.1\textwidth}}@{}}
        \shortstack{Object\\reference} & Background & GT & \shortstack{OmniPaint\\+ SFT} & \scalebox{0.88}{\shortstack{Insert-Anything\\+ SFT}} & SHINE & \shortstack{DIRECT\\+ GT Pose} & \shortstack{GPT-\\Image-2} & \shortstack{Nano Banana\\Pro} & \textbf{Ours} \\
    \end{tabular}
    \endgroup
    \par\vspace{2pt}
    \includegraphics[width=\linewidth]{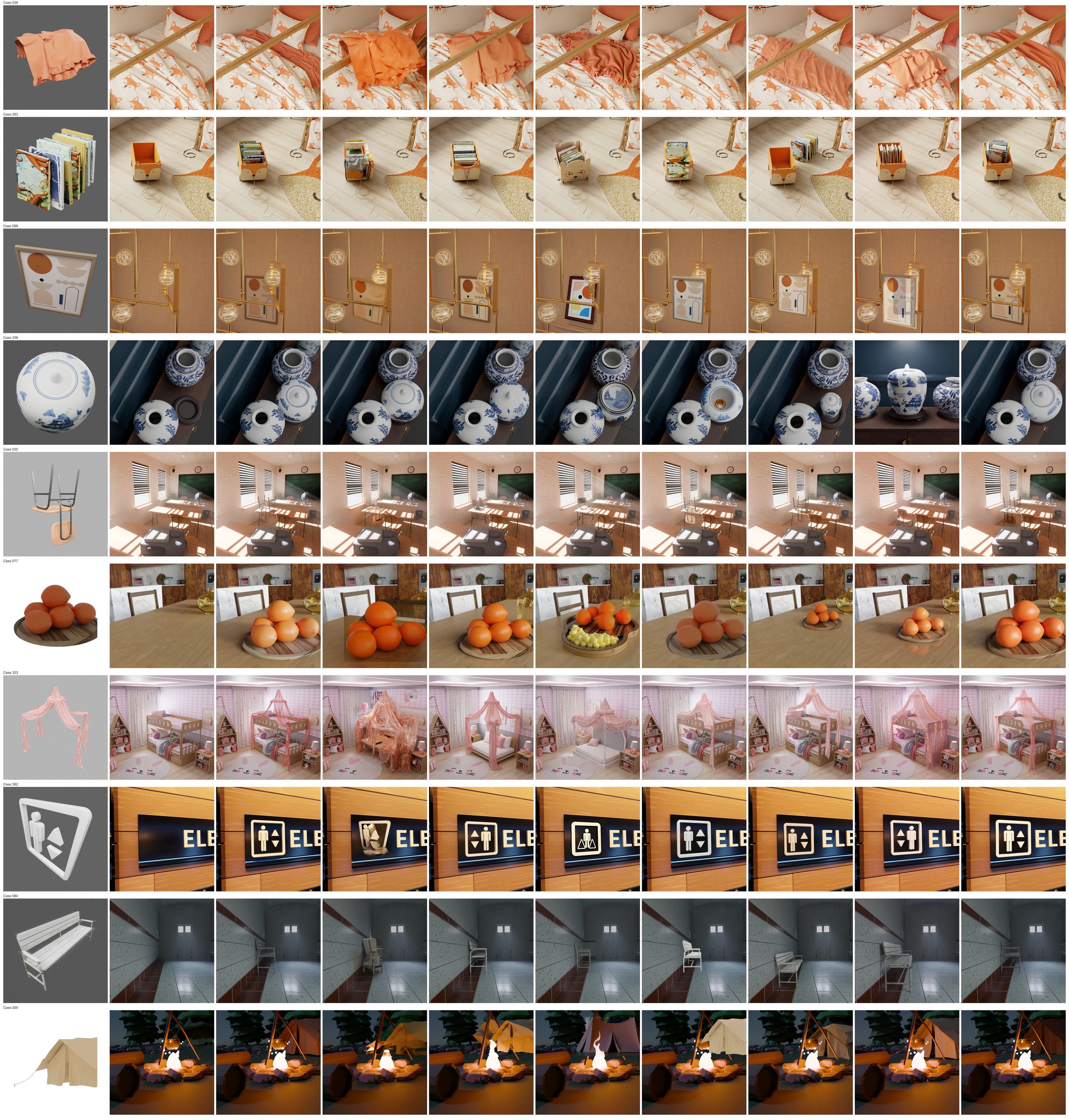}
    \caption{Selected box-guided comparisons on PhysInsert-Bench with one reference image. DIRECT additionally receives GT pose.}
    \label{fig:supp-physinsert-box-1view}
\end{figure}

\begin{figure}[!ht]
    \centering
    \begingroup
    \setlength{\tabcolsep}{0pt}
    \scriptsize
    \begin{tabular}{@{}*{9}{>{\centering\arraybackslash}p{0.111111\textwidth}}@{}}
        \shortstack{Object\\reference} & Background & GT & \shortstack{OmniPaint\\+ SFT} & \shortstack{Insert-Anything\\+ SFT} & JoyAI & \shortstack{GPT-\\Image-2} & \shortstack{Nano Banana\\Pro} & \textbf{Ours} \\
    \end{tabular}
    \endgroup
    \par\vspace{2pt}
    \includegraphics[width=\linewidth]{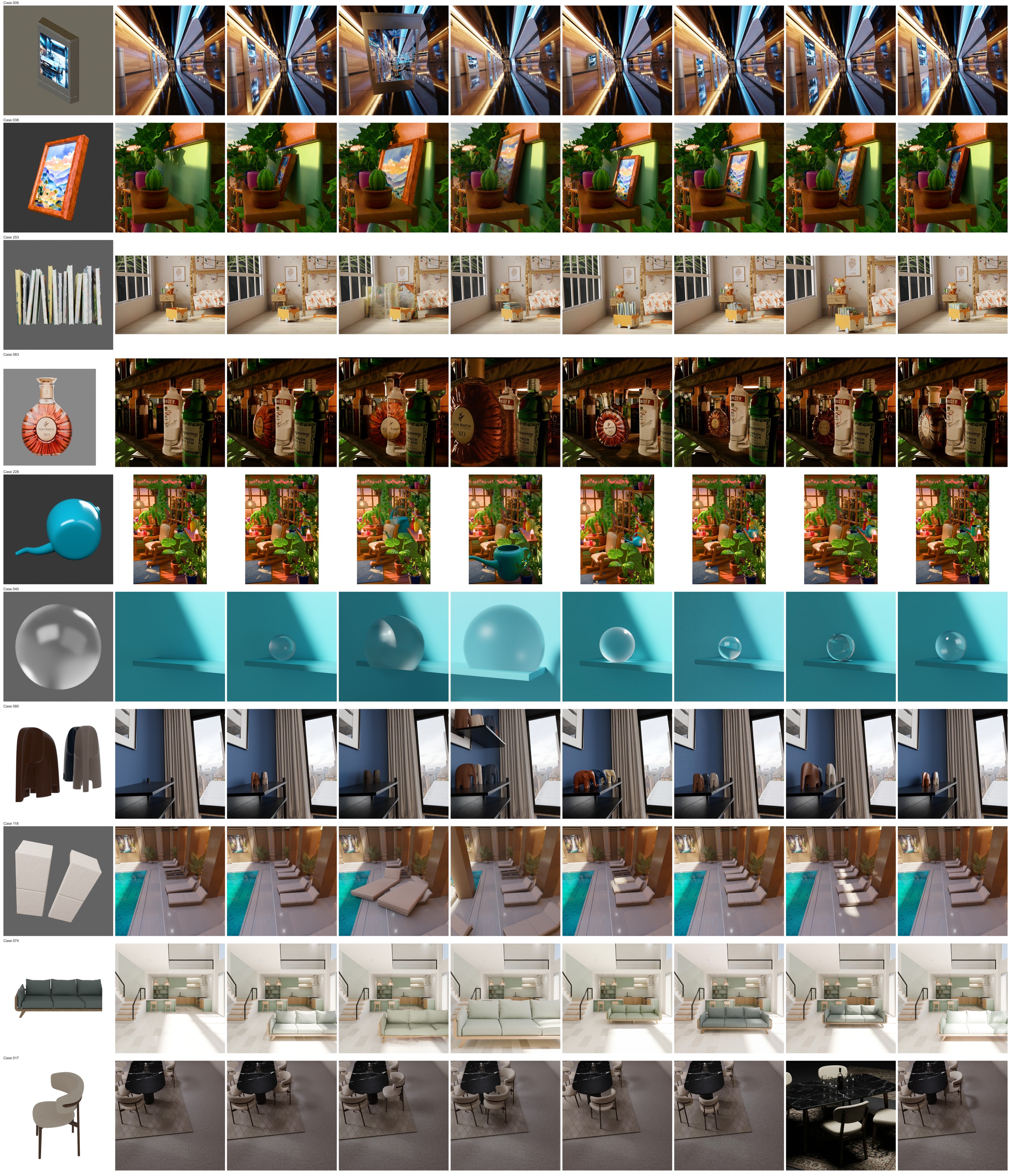}
    \caption{Selected text-guided comparisons on PhysInsert-Bench with one reference image. References are center-cropped only for visualization.}
    \label{fig:supp-physinsert-text-1view}
\end{figure}

Figures~\ref{fig:supp-physinsert-box-1view} and~\ref{fig:supp-physinsert-text-1view} show ten selected single-reference cases per guidance setting on PhysInsert-Bench, illustrating relative advantages in placement, viewpoint adaptation, and scene integration. Box-guided Ours uses the additional 4,000-step adaptation from Table~\ref{tab:box-guide}; text-guided Ours uses the text-guided checkpoint without this adaptation. Object references are center-cropped for display only; inference inputs and generated outputs are unchanged.

\FloatBarrier

\clearpage
\FloatBarrier

\begin{figure}[!ht]
    \centering
    \includegraphics[width=\linewidth]{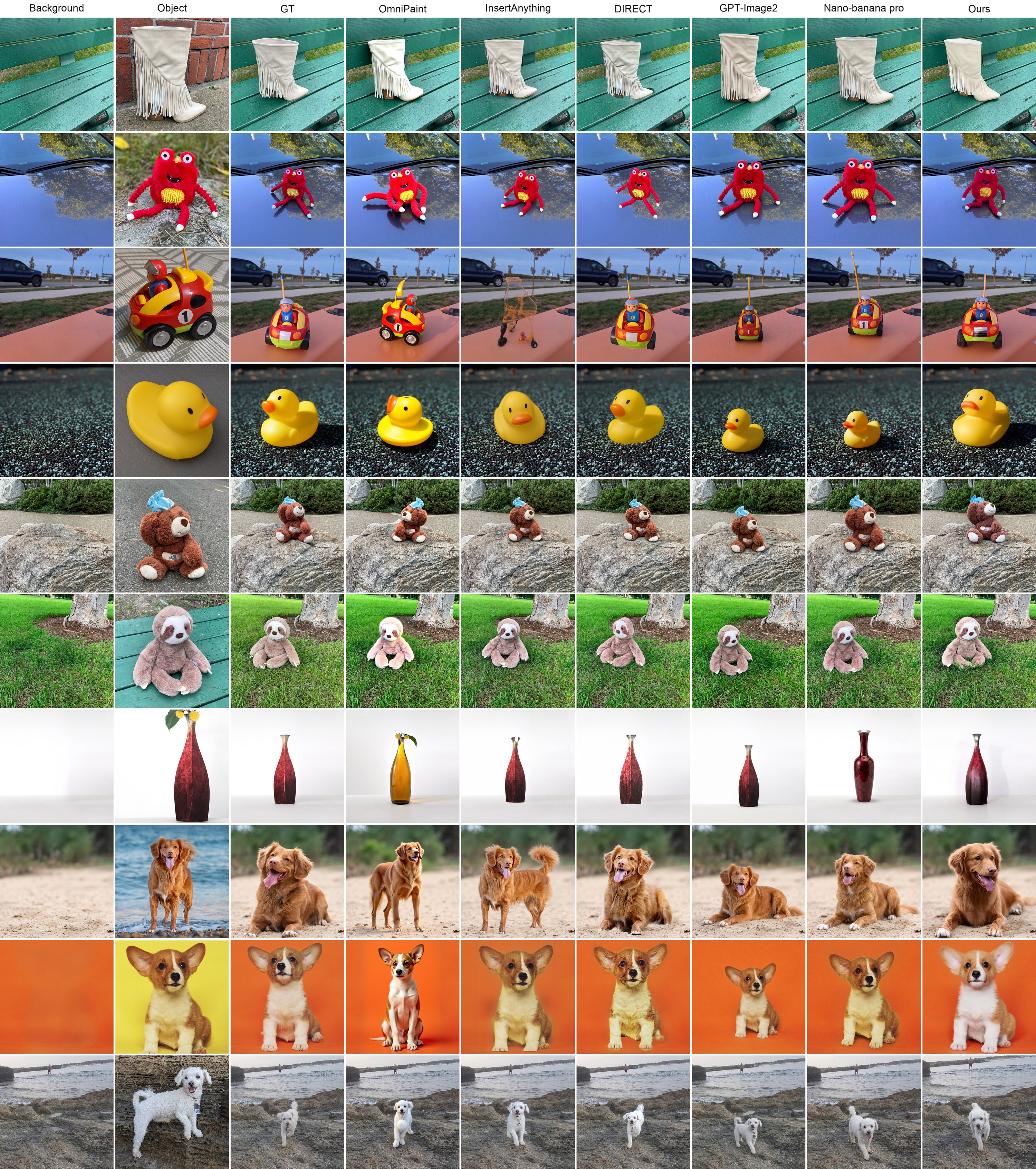}
    \caption{Additional qualitative comparisons on DreamBooth (1/2). Columns show the background, object reference, ground-truth target (GT), OmniPaint, InsertAnything, DIRECT, GPT-Image2, Nano Banana Pro, and Ours. DIRECT uses a GT-pose RGB proxy; GPT-Image2 and Nano Banana Pro use text guidance.}
    \label{fig:supp-dreambooth-1}
\end{figure}

\begin{figure}[!ht]
    \centering
    \includegraphics[width=\linewidth]{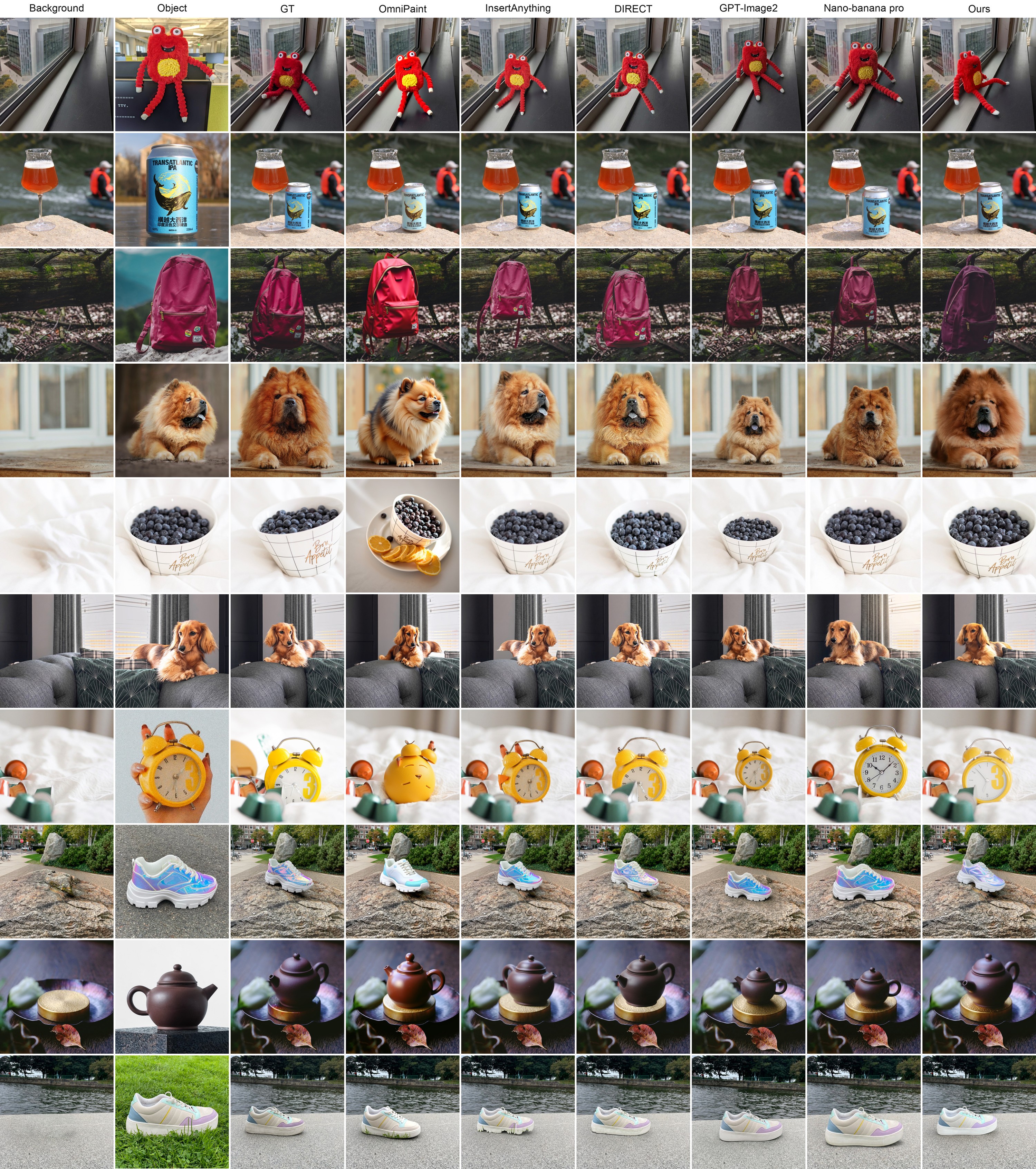}
    \caption{Additional qualitative comparisons on DreamBooth (2/2). The column order and method settings are the same as in Figure~\ref{fig:supp-dreambooth-1}.}
    \label{fig:supp-dreambooth-2}
\end{figure}

\begin{figure}[!ht]
    \centering
    \includegraphics[width=\linewidth]{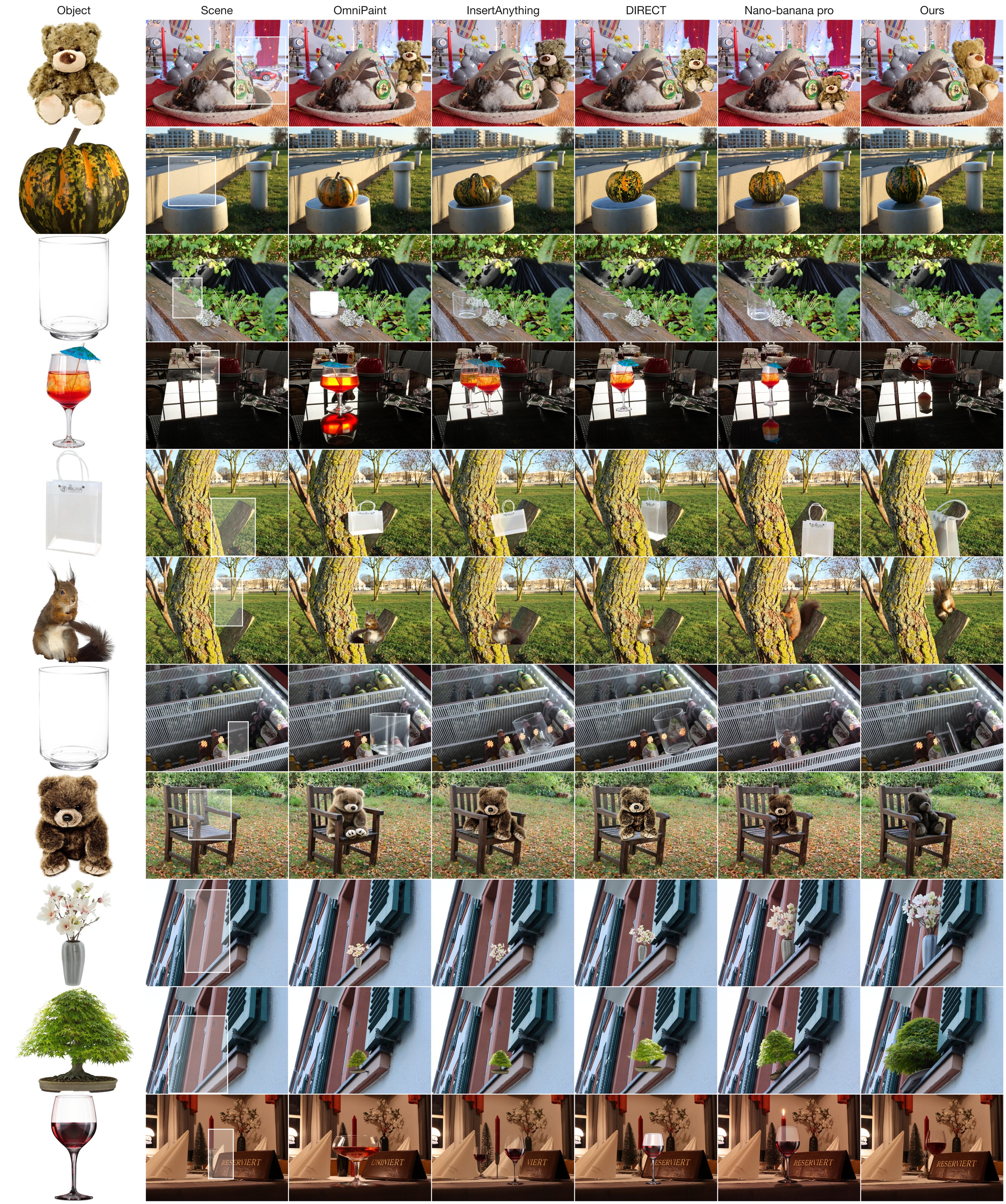}
    \caption{Additional qualitative comparisons on 11 cases from the in-the-wild user-study set. Columns show the object reference, background scene, OmniPaint, InsertAnything, DIRECT, Nano Banana Pro, and Ours. White boxes in the scene column indicate approximate insertion regions. OmniPaint is included for qualitative comparison only and was not part of the user-study ratings.}
    \label{fig:supp-user-study}
\end{figure}

\clearpage
Figures~\ref{fig:supp-dreambooth-1} and~\ref{fig:supp-dreambooth-2} present additional qualitative comparisons on 20 DreamBooth cases, with one reference image per case. Figure~\ref{fig:supp-user-study} presents further comparisons on 11 cases from the separate in-the-wild user-study set.

\clearpage
\FloatBarrier

\clearpage
\FloatBarrier

\clearpage
\section{Dataset Construction Details}
\label{app:dataset-construction}

\paragraph{Sample Filtering and Deduplication.}
We randomly select candidate objects from complete BlenderKit scenes and render counterfactual pairs under fixed camera, lighting, and remaining scene settings, recomputing shadows, mirror images, and reflections after removal. Rule-based filtering first rejects pairs with excessively small or large image changes. Gemini then examines the scene pair and object views for object completeness, background plausibility after removal, image quality, and realism, excluding unsuitable scene structures, residual semantic parts, and visible rendering artifacts. We deduplicate retained objects using DINO feature similarity while retaining their associated scenes.
The complete Gemini quality-scoring and interaction-classification prompts, including image order and JSON output examples, are reproduced in Appendix~\ref{app:gemini-curation-prompts}.

\paragraph{Scene Augmentation.}
For each retained scene--object combination, we keep the object and scene configuration unchanged while randomly varying the camera position. At each viewpoint, we rerender a matched with/without pair and apply a second quality check.

\paragraph{Object Reference Rendering.}
Each deduplicated object is independently rendered under seven lighting conditions: fixed white light, fixed studio lighting, and five randomly sampled HDR environments. For each condition, we use 11 fixed and 12 random viewpoints while preserving object geometry and material, yielding $7\times23=161$ reference images per object. These references are associated with the corresponding scene pairs.

\paragraph{Dataset Comparison.}
Table~\ref{tab:dataset-comparison} compares AnyInsertion~\citep{InsertAnything}, DIRECT~\citep{Direct}, and InsertWorld. CF denotes object-presence/absence pairs acquired through controlled capture or physical re-rendering under otherwise fixed settings; MV/ML denote same-object multiview/multilight data, and Ctrl. denotes control inputs. A blue check mark indicates that the corresponding data is provided; a red cross indicates that it is not specifically constructed; Partial indicates partial coverage.

\subsection{Controlled Benchmark Construction}
\label{app:benchmark-construction}

Each controlled PhysInsert-Bench example contains an object-free background $I_{-}$, a reference set $\mathcal{I}_r$, an insertion instruction $p$, the corresponding with-object target $I_{+}$, and the target object region. The target box is the 2D bounding box enclosing the object rendered in Blender at its target pose. For box guidance, an additional reference image annotated with this box is supplied as visual conditioning, alongside the object-reference images. The object-reference counts reported in the tables do not include this additional box-annotated image. Text guidance instead uses spatial instructions describing position, scale, orientation, and occlusion relationships without an inference-time mask.

Each case also carries physical-effect labels and evaluation-focus tags, which specify the interactions emphasized by that example and are incorporated into the fixed Phy-VLM prompt. Interaction annotations describe scene-to-object requirements in illumination, visibility, and placement, and object-to-scene effects including cast shadows, reflections, mirror images, and occlusion changes. Optical cases combine substantial reference--target illumination differences with visible environmental effects. Spatial \& Mechanical cases and 26 Optical challenges provide independent reference viewpoints to test scene-dependent orientation or pose adaptation.

\subsection{Real-World Test Set Construction}
\label{app:real-world-test-construction}

\paragraph{DreamBooth.}
We construct 40 test cases covering all objects in DreamBooth. For each case, we randomly select one image as the target and use the remaining images of the same object as the reference pool; we report one-reference evaluation. We remove the object from the target image using Nano Banana Pro and manually review the resulting background.

\paragraph{PhysInsert-Bench-Wild.}
The 50 cases draw on Amazon Berkeley Objects (ABO)~\citep{ABO}, NAVI v1.5~\citep{NAVI}, the training and validation splits of ORIDa~\citep{ORIDa}, and real images with references generated using SAM 3D Objects~\citep{SAM3DObjects}. Reference pools contain multiview product images from ABO, multiview photographs of the same object from NAVI, and cross-scene observations of the same object from ORIDa. For the SAM 3D-based cases, we use real target images and render reference views from SAM 3D Objects reconstructions.

\clearpage
\section{Dataset and Benchmark Samples}
\label{app:dataset-benchmark-samples}

We visualize five examples from the InsertWorld training set and twenty examples from PhysInsert-Bench. The training examples show counterfactual scene pairs, isolated objects, and references under different viewpoints and illumination. The benchmark examples cover twenty distinct interaction tags, with one example per selected tag. These are illustrative data samples, not model predictions or a performance-ranked selection. Images retain their original aspect ratios and are resized for display. In the training examples, the original transparency of Removed-only renders is displayed over white; all other image contents are unchanged.

\begin{figure}[!ht]
    \centering
    \includegraphics[width=\linewidth,height=0.83\textheight,keepaspectratio]{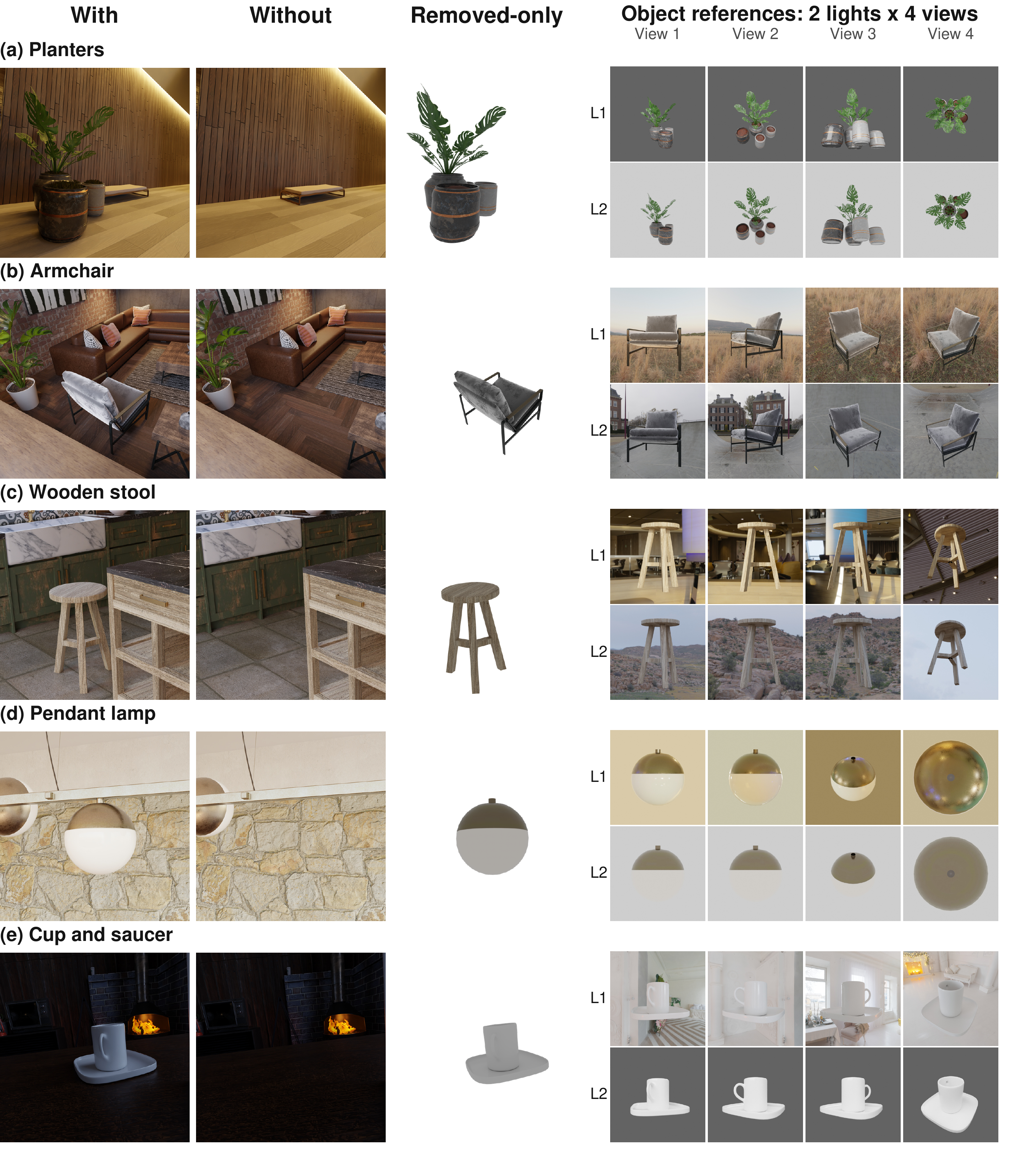}
    \caption{Five InsertWorld training examples. Each row contains the with-object scene (With), the corresponding object-free scene (Without), and the isolated target rendered from the scene viewpoint on a white display background (Removed-only). The right panel shows the same target under two lighting conditions (rows L1 and L2) and four corresponding fixed viewpoints (columns). Lighting conditions and viewpoints are selected separately for each example.}
    \label{fig:supp-training-samples}
\end{figure}
\FloatBarrier

\clearpage
\begin{figure}[!ht]
    \centering
    \includegraphics[width=\linewidth,height=0.83\textheight,keepaspectratio]{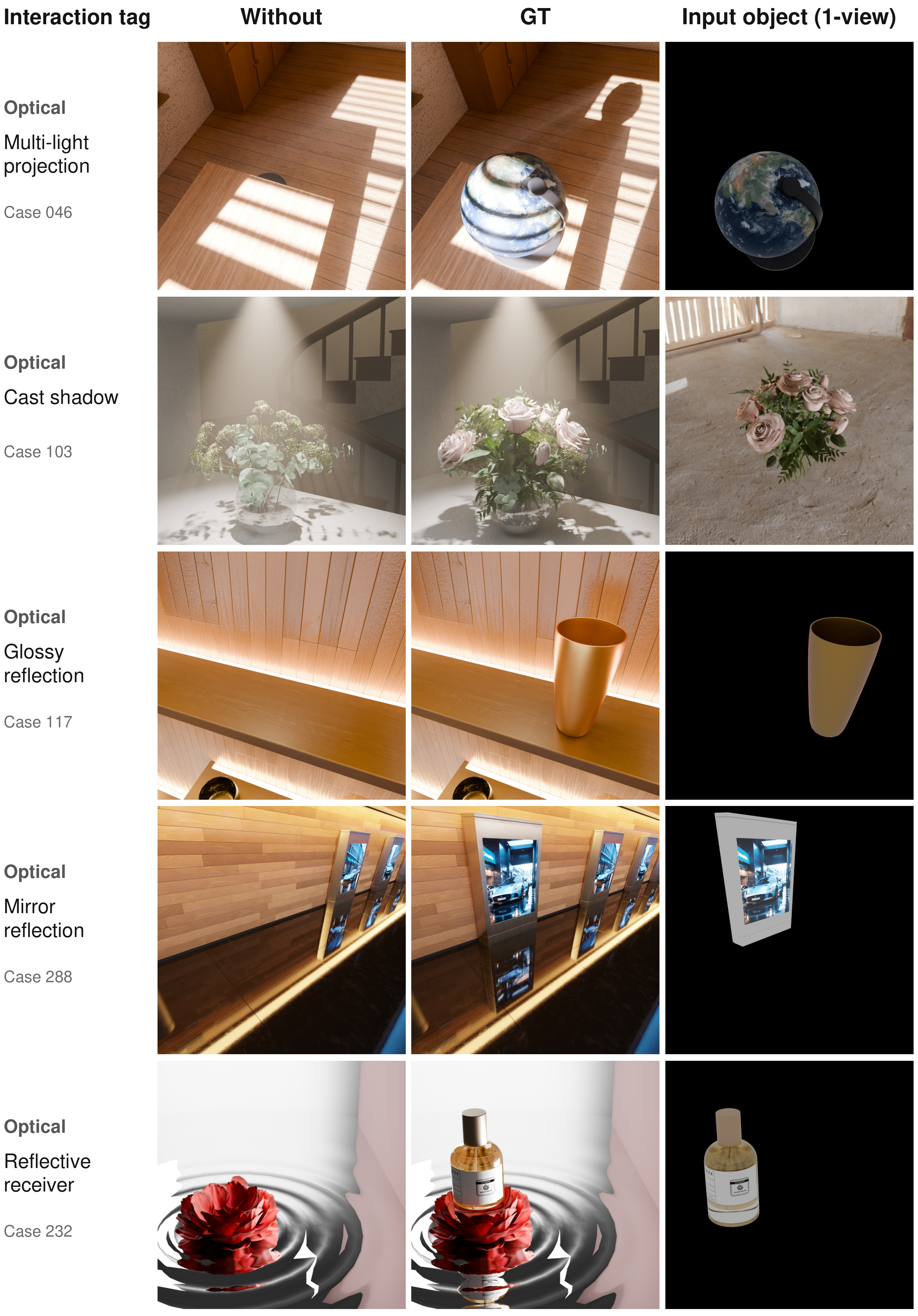}
    \caption{PhysInsert-Bench samples (1/4). The original interaction tag appears at left in readable form, followed by the object-free scene (Without), with-object target (GT), and actual single-view object input. Case IDs refer to the benchmark manifest.}
    \label{fig:supp-benchmark-samples-1}
\end{figure}
\FloatBarrier

\clearpage
\begin{figure}[!ht]
    \centering
    \includegraphics[width=\linewidth,height=0.83\textheight,keepaspectratio]{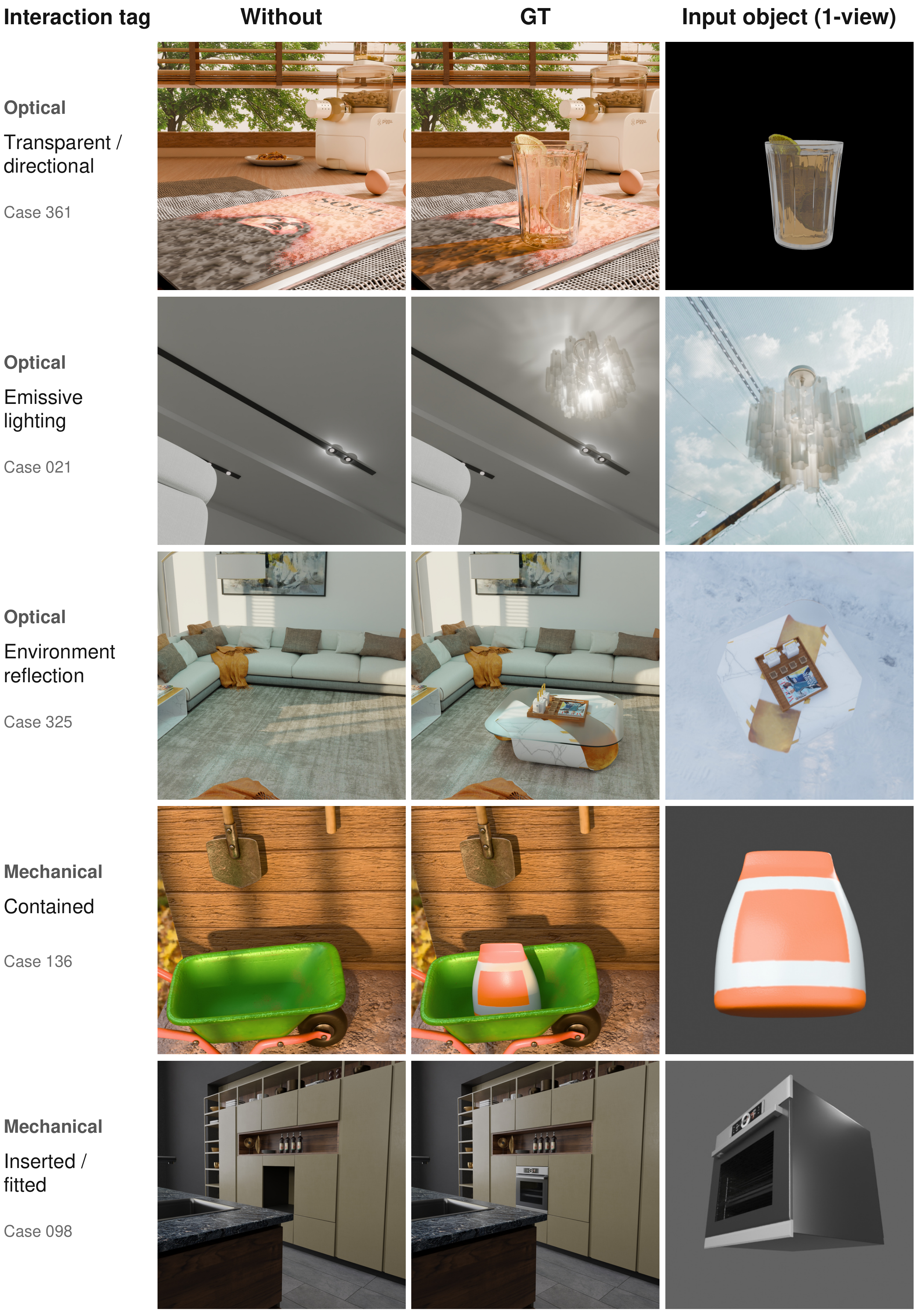}
    \caption{PhysInsert-Bench samples (2/4). Column meanings follow Figure~\ref{fig:supp-benchmark-samples-1}. The reference is the actual one-view benchmark input, including its original background; it is neither a multiview collage nor a newly rendered reference.}
    \label{fig:supp-benchmark-samples-2}
\end{figure}
\FloatBarrier

\clearpage
\begin{figure}[!ht]
    \centering
    \includegraphics[width=\linewidth,height=0.83\textheight,keepaspectratio]{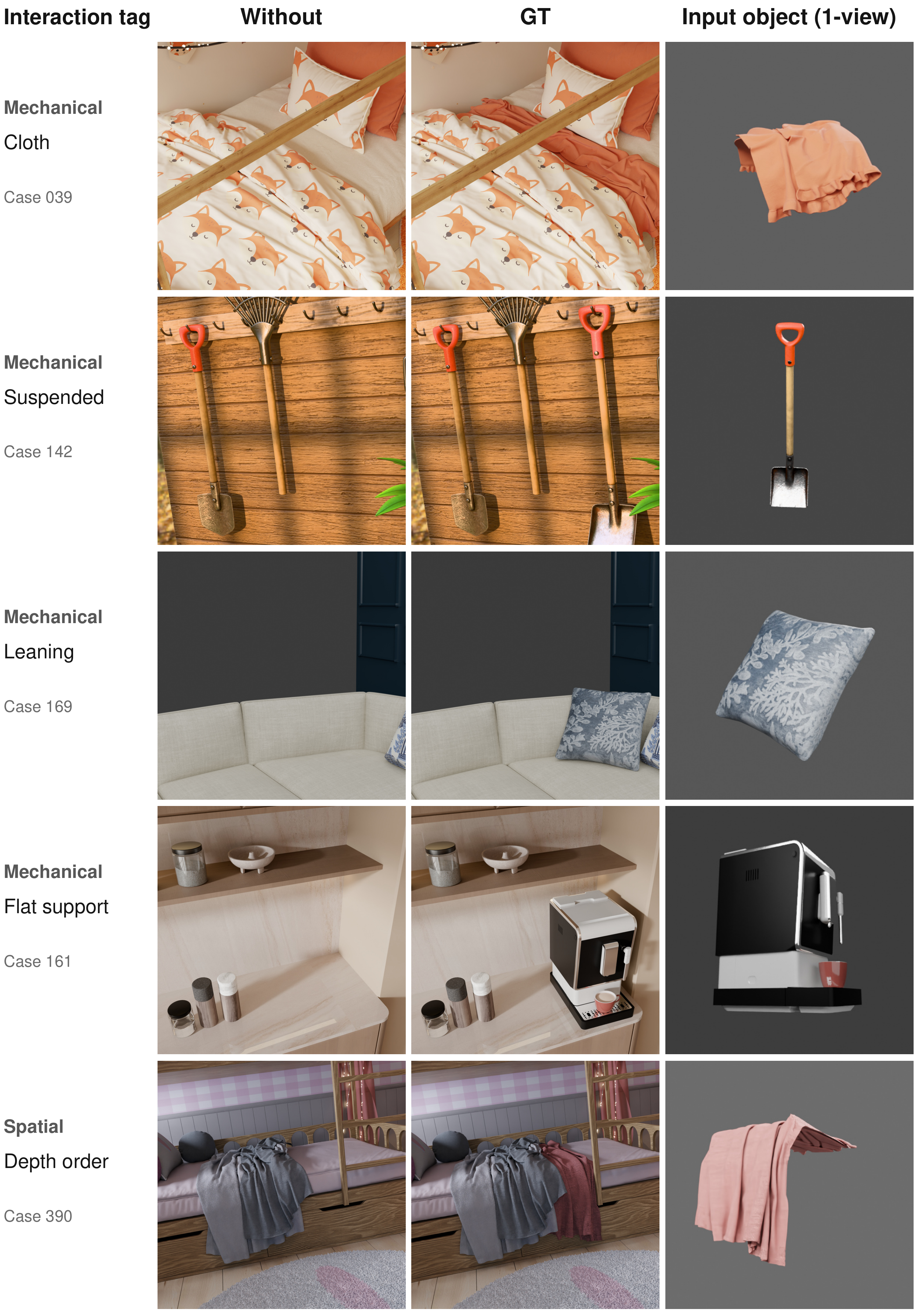}
    \caption{PhysInsert-Bench samples (3/4). Column meanings follow Figure~\ref{fig:supp-benchmark-samples-1}. The examples illustrate cloth, suspension, leaning, flat support, and depth ordering using the original benchmark annotations.}
    \label{fig:supp-benchmark-samples-3}
\end{figure}
\FloatBarrier

\clearpage
\begin{figure}[!ht]
    \centering
    \includegraphics[width=\linewidth,height=0.83\textheight,keepaspectratio]{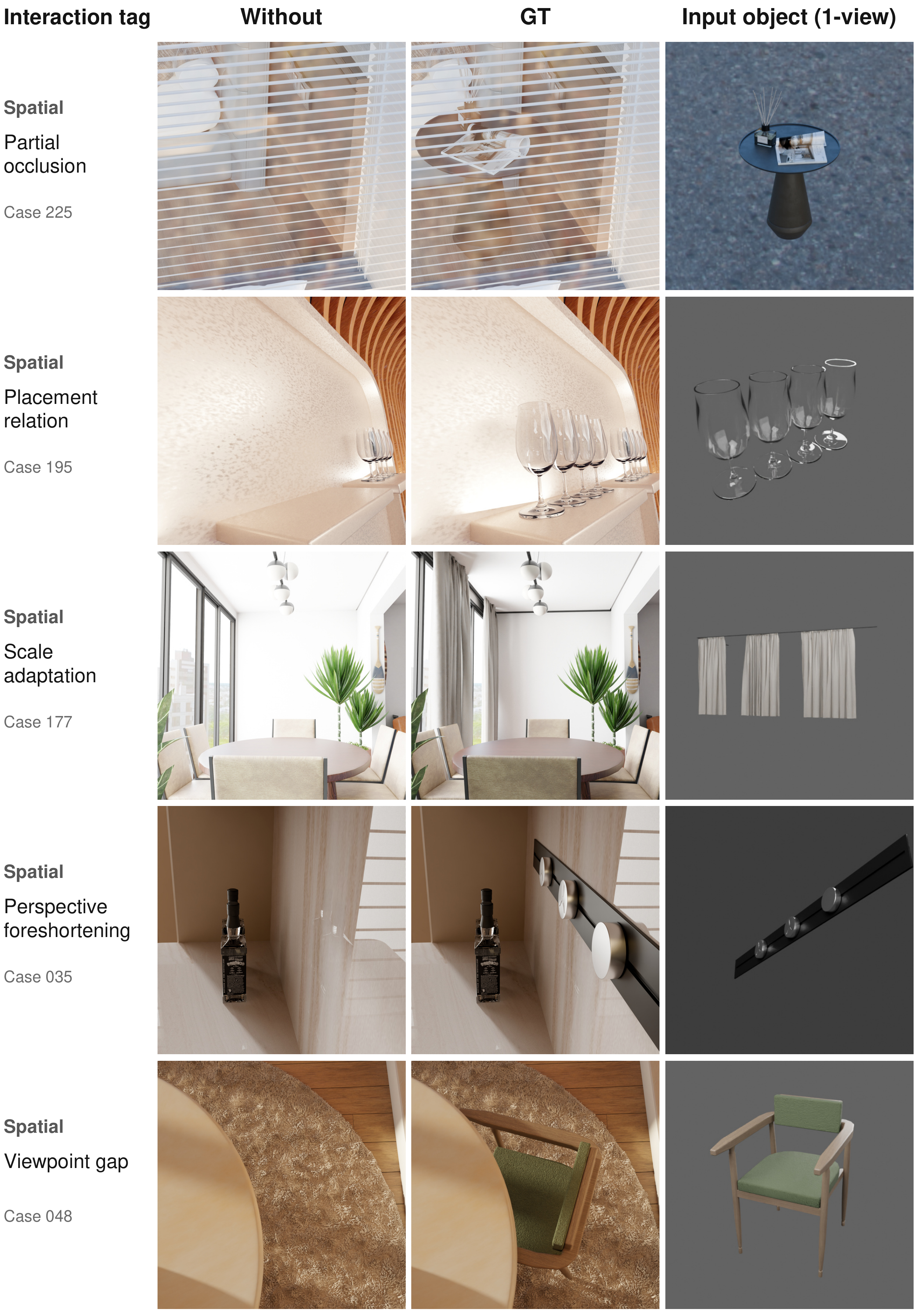}
    \caption{PhysInsert-Bench samples (4/4). Column meanings follow Figure~\ref{fig:supp-benchmark-samples-1}. These examples cover partial occlusion, placement relations, scale adaptation, perspective foreshortening, and viewpoint gaps.}
    \label{fig:supp-benchmark-samples-4}
\end{figure}
\FloatBarrier

\clearpage
\clearpage
\subsection{Gemini Prompts for Dataset Curation}
\label{app:gemini-curation-prompts}

We reproduce the quality-scoring and interaction-classification prompts used by the dataset-curation implementation. These prompts inspect rendered training-data candidates, not predictions from the methods compared in our result tables. Both produce a short insertion caption (\texttt{insertion\_description}) as part of a structured JSON response; this caption is distinct from the full editing prompt used to condition the generation model. The wording and example responses below are preserved, with only typographic reformatting. The JSON examples are part of the prompts, not measured sample annotations.

\paragraph{Ordered visual inputs.}
Both stages receive the same first six images. Classification additionally receives an environment-difference visualization. A single prompt accompanies the ordered image list; changing this order would change the intended interpretation.

\begin{center}
\small
\setlength{\tabcolsep}{6pt}
\renewcommand{\arraystretch}{1.13}
\begin{tabular}{@{}clp{0.56\linewidth}@{}}
\toprule
\textbf{Index} & \textbf{Input} & \textbf{Role} \\
\midrule
1 & \texttt{with} & Scene with the candidate target inserted. \\
2 & \texttt{without} & Matched scene after target removal. \\
3 & \texttt{removed\_only} & Isolated target/group in the scene view. \\
4--6 & Front, right, top & Three canonical views of the target/group. \\
7 & Environment difference & Classification only: signed changes outside the target; blue/darker, red/brighter, green/chromatic. \\
\bottomrule
\end{tabular}
\end{center}

\paragraph{Two complementary annotations.}
Stage 1 assigns six integer scores in $[0,5]$: object validity, three-view quality, insertion-result validity, image realism, insertion difficulty, and overall quality. Difficulty rewards useful complexity but must not compensate for severe quality failures. Stage 2 retains an image-realism score, assigns relative scale and mechanical type, and records seven scene-interaction attributes. Mechanical interaction and lighting/shadow complexity use integer levels $0/1/2$; the other five interaction fields are Boolean. These are evidence-based annotations, not mutually exclusive benchmark-subset labels. In particular, a visible change in the amplified difference map is not sufficient evidence of a reflection without confirmation in the original scene pair.

\paragraph{Version scope.}
The following templates correspond to the six-image quality-scoring and seven-image interaction-classification implementations. Older prompt variants and batch-level model overrides may differ; reproducing a historical annotation also requires that batch's recorded request configuration. The templates do not by themselves specify a universal dataset-retention threshold.

\begingroup
\tcbset{curationprompt/.style={
  enhanced,breakable,
  colback=black!2,colframe=black!28,
  colbacktitle=black!7,coltitle=black,
  boxrule=0.4pt,arc=1.3mm,
  left=9pt,right=9pt,top=7pt,bottom=7pt,
  before skip=10pt,after skip=10pt,
  fonttitle=\small\bfseries,
  fontupper=\fontsize{9}{11}\selectfont,
  pad at break*=2mm,
  title after break={\small\bfseries\insertcontinuationtext}
}}
\emergencystretch=2em

\subsubsection{Stage 1: Six-Dimension Quality Scoring}
\label{app:gemini-quality-prompt}
\def\insertcontinuationtext{Prompt S1: Quality scoring (continued)}
\begin{tcolorbox}[curationprompt,title={Prompt S1: Quality scoring --- complete request text}]
\noindent You are a senior quality-control expert for 3D object-insertion training data. You will receive 6 images and must judge whether this sample is suitable for object-insertion training.\par
\smallskip
\par\medskip\noindent\textbf{[Inputs]}\par\nobreak
\noindent\hangindent=1em\hangafter=1 - Image 1: with, the result after inserting the object.\par
\noindent\hangindent=1em\hangafter=1 - Image 2: without, the background before insertion.\par
\noindent\hangindent=1em\hangafter=1 - Image 3: removed\_\allowbreak{}only, an isolated rendering of the object to be inserted.\par
\noindent\hangindent=1em\hangafter=1 - Image 4: front view of the object.\par
\noindent\hangindent=1em\hangafter=1 - Image 5: right view of the object.\par
\noindent\hangindent=1em\hangafter=1 - Image 6: top view of the object.\par
\smallskip
\par\medskip\noindent\textbf{[Task]}\par\nobreak
\noindent 1. Describe the sample from the insertion perspective: "insert [object] at [location]". Treat Image 1 as the result of inserting the object shown in Images 3-6 into the scene in Image 2. If the object cannot be identified, write "unclear insertion object".\par
\noindent 2. Score the following 6 dimensions from 0 to 5, where 5 is best. Use Images 3-6 especially for object\_\allowbreak{}validity and three\_\allowbreak{}view\_\allowbreak{}quality, and compare Images 1-2 especially for insertion\_\allowbreak{}result\_\allowbreak{}validity.\par
\noindent 3. Output JSON keys must be English exactly as specified. Chinese labels below are only references.\par
\smallskip
\noindent If the inserted object is unclear, insertion\_\allowbreak{}description must be "unclear insertion object", and all six scores must be 0.\par
\smallskip
\par\medskip\noindent\textbf{[Dimensions]}\par\nobreak
\par\medskip\noindent\textbf{1. object\_\allowbreak{}validity}\par\nobreak
\noindent\hangindent=1em\hangafter=1 - Judge whether the target is a reasonable independent object or a tightly bound object group.\par
\noindent\hangindent=1em\hangafter=1 - Reasonable examples: cups, chairs, flowerpots, lamps, curtains, wire mesh, hanging ornaments, decorative objects, or strongly bound small parts.\par
\noindent\hangindent=1em\hangafter=1 - Unreasonable examples: walls, floors, ceilings, large background regions, large cabinet/room structures, or structural surfaces that change the whole scene.\par
\noindent\hangindent=1em\hangafter=1 - If removed\_\allowbreak{}only or the three views show a wall, floor, ceiling, or huge background structure, score \textless{}= 1.\par
\noindent\hangindent=1em\hangafter=1 - Curtains, rugs, wire mesh, railings, hanging screens, etc. may be large but can still be valid if they are semantically independent insertable objects.\par
\noindent\hangindent=1em\hangafter=1 - Thin disk-like or sheet-like objects such as coasters, pads, washers, thin trays, dials, rugs, and similar items are valid when they have clear semantics or physical relations, such as a cup-coaster support/contact relation. Do not give a low score only because they are thin.\par
\noindent\hangindent=1em\hangafter=1 - The inserted target should preferably be one object or a semantically coherent object group. Spatial separation is acceptable when the objects form one continuous semantic unit or a natural arrangement, such as a row of chairs. However, if the target contains multiple disconnected object groups placed in different unrelated regions, such as several groups of books in different compartments of a bookshelf, score \textless{}= 2.\par
\smallskip
\par\medskip\noindent\textbf{2. three\_\allowbreak{}view\_\allowbreak{}quality}\par\nobreak
\noindent\hangindent=1em\hangafter=1 - Judge whether Images 4-6 provide enough clear, learnable information about the inserted object.\par
\noindent\hangindent=1em\hangafter=1 - At least one or more views should clearly show the object's outline, scale, orientation, part relationship, or key material information. Penalize if all three views are blurry, overexposed, too dark, too small, heavily cropped, or nearly duplicate the same view.\par
\noindent\hangindent=1em\hangafter=1 - Do not give a low score just because one or two views look like a line. Flat objects such as disks, coasters, pads, thin boards, trays, and dials naturally look line-like from front/side views. If the top view or another view clearly shows the shape, and the object has clear semantics or a physical relation, this dimension can be medium or high.\par
\noindent\hangindent=1em\hangafter=1 - If at least one three-view image provides enough information, for example a coaster/disk/dial shape is clear in top view, the score should usually be \textgreater{}= 3.\par
\noindent\hangindent=1em\hangafter=1 - Score \textless{}= 2 only when all three views provide little useful information, or when the object is extremely thin/long and lacks learnable shape, material, or semantic information, such as a very thin railing/rod/cable.\par
\smallskip
\par\medskip\noindent\textbf{3. insertion\_\allowbreak{}result\_\allowbreak{}validity}\par\nobreak
\noindent\hangindent=1em\hangafter=1 - Focus on Image 2 without: whether it is suitable as the pre-insertion background. The sample was constructed by deletion, so with is usually more reliable and without is more likely to contain artifacts.\par
\noindent\hangindent=1em\hangafter=1 - Core check: whether Image 2 still contains parts that belong to the same semantic object, same object group, or strongly bound components of the inserted object. For a flowerpot, without should not leave leaves/soil/tray from the same plant; for a pendant lamp, without should not leave the lampshade/bulb/wire/rod of the same lamp.\par
\noindent\hangindent=1em\hangafter=1 - If unrelated objects in without are floating, structurally odd, abstract, or physically unnatural, but they are not the same semantic object/group and not strongly bound to the inserted object, do not heavily penalize this dimension.\par
\noindent\hangindent=1em\hangafter=1 - Penalize this dimension only when without has same-object residuals, same-group residuals, strongly bound residual parts, or an obvious hole/break at the insertion region that makes it unusable as a pre-insertion background.\par
\noindent\hangindent=1em\hangafter=1 - If Image 1 contains an obvious violation of physical plausibility in an otherwise highly realistic scene, such as a clearly floating object in an indoor environment, or if the inserted object has physically implausible mesh penetration with the scene, the score for this dimension should be \textless{}= 2.\par
\noindent\hangindent=1em\hangafter=1 - If the difference between the Image 1 and Image 2 is too small, such as when the inserted object is heavily occluded or visually too similar to the surrounding environment, making the insertion result difficult to identify, the score for this dimension should be \textless{}= 2.\par
\smallskip
\par\medskip\noindent\textbf{4. image\_\allowbreak{}realism}\par\nobreak
\noindent\hangindent=1em\hangafter=1 - Judge whether Images 1 and 2 contain rendering or geometry quality problems that make the sample unusable. Do not simply penalize the scene for being stylized or abstract.\par
\noindent\hangindent=1em\hangafter=1 - Acceptable: minimal abstract geometry scenes, simple 3D style, unrelated floating objects in without, unrelated physically unnatural context, or small black rendering artifacts outside the object insertion region that do not affect insertion quality.\par
\noindent\hangindent=1em\hangafter=1 - Accept reasonable contact, support, attachment, or slight embedding between an object and its support, such as a cup on a coaster or an object attached to a pad/base. Do not label such physical contact as mesh penetration.\par
\noindent\hangindent=1em\hangafter=1 - Penalize: severe visible mesh penetration, broken geometry, abnormal faces, wrong-normal black/bright faces, obviously wrong material rendering, abnormal exposure or lighting, large black/blank image regions, large unintended solid-color backgrounds, or rendering collapse.\par
\noindent\hangindent=1em\hangafter=1 - If the image contains large black unrendered regions, or if the scene lighting is so low that key scene information is difficult to recognize, the score should be \textless{}= 2.\par
\smallskip
\par\medskip\noindent\textbf{5. insertion\_\allowbreak{}difficulty}\par\nobreak
\noindent\hangindent=1em\hangafter=1 - This is a bonus-like dimension. It measures whether the sample contains difficult but valuable insertion conditions. It must not hide severe failures in the basic dimensions above.\par
\noindent\hangindent=1em\hangafter=1 - Difficulty has three categories: spatial complexity (occlusion, perspective deformation), physical complexity (size, support, tilt, suspension, contact), and lighting/material complexity (high contrast, strong shadows, reflection, self-emitting lamp-like objects, metal/transparent/glass/complex textures).\par
\noindent\hangindent=1em\hangafter=1 - If the object is frontal, unoccluded, under simple lighting, with simple material and simple physical relation, difficulty should be low or medium.\par
\smallskip
\par\medskip\noindent\textbf{6. overall\_\allowbreak{}quality}\par\nobreak
\noindent\hangindent=1em\hangafter=1 - Overall suitability for object-insertion training, considering the five dimensions above.\par
\noindent\hangindent=1em\hangafter=1 - If any basic dimension among object\_\allowbreak{}validity, three\_\allowbreak{}view\_\allowbreak{}quality, insertion\_\allowbreak{}result\_\allowbreak{}validity, and image\_\allowbreak{}realism has a severe error, overall\_\allowbreak{}quality must be low and cannot be rescued by high insertion\_\allowbreak{}difficulty.\par
\noindent\hangindent=1em\hangafter=1 - If the four basic dimensions have no severe problem, insertion\_\allowbreak{}difficulty can boost overall\_\allowbreak{}quality. High-quality high-difficulty samples can receive 5. Low-difficulty samples usually tend toward 3-4 unless the basic dimensions are exceptionally good.\par
\smallskip
\par\medskip\noindent\textbf{[Scoring Discipline]}\par\nobreak
\noindent Be strict. Do not give 4 or 5 lightly; reserve high scores for samples with clearly strong evidence and no serious issues in the relevant dimensions.\par
\noindent For each dimension, the reason must be specific: point out the problem location and problem type, and avoid vague generic comments.\par
\smallskip
\par\medskip\noindent\textbf{[Output Requirements]}\par\nobreak
\noindent Return only one JSON object wrapped by ```json and ```. Do not output any extra text.\par
\noindent The JSON object must contain exactly these top-level keys: insertion\_\allowbreak{}description, object\_\allowbreak{}validity, three\_\allowbreak{}view\_\allowbreak{}quality, insertion\_\allowbreak{}result\_\allowbreak{}validity, image\_\allowbreak{}realism, insertion\_\allowbreak{}difficulty, overall\_\allowbreak{}quality, reason.\par
\noindent reason must be a JSON object, not a string. It must contain exactly these six keys: object\_\allowbreak{}validity, three\_\allowbreak{}view\_\allowbreak{}quality, insertion\_\allowbreak{}result\_\allowbreak{}validity, image\_\allowbreak{}realism, insertion\_\allowbreak{}difficulty, overall\_\allowbreak{}quality.\par
\noindent All six score values must be integers from 0 to 5.\par
\smallskip
\par\medskip\noindent\textbf{[Valid Output Template]}\par\nobreak
\begin{Verbatim}[fontsize=\fontsize{8.3}{10}\selectfont,breaklines=true,breakanywhere=true,breaksymbolleft={},breaksymbolright={}]
{
  "insertion_description": "insert a white vase on the tabletop",
  "object_validity": 5,
  "three_view_quality": 4,
  "insertion_result_validity": 4,
  "image_realism": 4,
  "insertion_difficulty": 2,
  "overall_quality": 4,
  "reason": {
    "object_validity": "The target is an independent vase and is a reasonable insertable object.",
    "three_view_quality": "The three views clearly show the vase outline, volume, and material cues.",
    "insertion_result_validity": "The without image is usable and has no same-object residual parts around the insertion region.",
    "image_realism": "The scene has no severe rendering collapse, broken geometry, or obvious material errors.",
    "insertion_difficulty": "The object is mostly unoccluded with simple lighting and material, so the insertion difficulty is moderate to low.",
    "overall_quality": "The four basic dimensions are acceptable, but the task is not very difficult, so it is a usable regular training sample."
  }
}
\end{Verbatim}
\end{tcolorbox}

\subsubsection{Stage 2: Object Properties and Scene Interactions}
\label{app:gemini-classification-prompt}
\def\insertcontinuationtext{Prompt S2: Interaction classification (continued)}
\begin{tcolorbox}[curationprompt,title={Prompt S2: Interaction classification --- complete request text}]
\noindent You are a senior analyst of 3D object-insertion data. You will receive seven images describing one insertion case. Classify the inserted object and its interactions with the scene. Also retain an image-realism quality score.\par
\smallskip
\par\medskip\noindent\textbf{[Input images]}\par\nobreak
\noindent\hangindent=1em\hangafter=1 - Image 1: with, the scene containing the target object.\par
\noindent\hangindent=1em\hangafter=1 - Image 2: without, the same scene after removing the target object.\par
\noindent\hangindent=1em\hangafter=1 - Image 3: removed\_\allowbreak{}only, an isolated rendering of the complete removed object group.\par
\noindent\hangindent=1em\hangafter=1 - Image 4: front view of the complete removed object group.\par
\noindent\hangindent=1em\hangafter=1 - Image 5: right view of the complete removed object group.\par
\noindent\hangindent=1em\hangafter=1 - Image 6: top view of the complete removed object group.\par
\noindent\hangindent=1em\hangafter=1 - Image 7: signed environment difference map after masking the target object's own pixels. Black means little or no detected change. Blue means the pre-existing scene became darker after insertion, red means it became brighter, and green means a mainly chromatic change with little luminance change. Intensity represents amplified change magnitude. Use it only to locate and type candidate changes, then verify their physical meaning in Images 1-2.\par
\smallskip
\noindent Use Images 3-6 to identify intrinsic object properties. Compare Images 1 and 2 to identify object-environment relations. Do not infer a relation merely because it is theoretically possible. For boolean fields, mark true only with visible evidence. For 0/1/2 fields, choose a positive level only with visible evidence supporting that level.\par
\smallskip
\par\medskip\noindent\textbf{[Tasks]}\par\nobreak
\noindent 1. Write insertion\_\allowbreak{}description in the form "insert [object] at [location]".\par
\noindent 2. Score image\_\allowbreak{}realism from 0 to 5, where 5 is best. The inserted target must be correctly rendered in Image 1 (with). Check Images 1 and 2 for rendering collapse, broken or missing geometry, wrong normals, severe penetration, abnormal exposure, missing textures, genuinely unrendered regions, abnormal black target parts, and clearly incorrect material rendering. Stylized but internally consistent scenes are acceptable.\par
\noindent\hangindent=1.7em\hangafter=1 - Judge rendering and geometry quality only to determine whether the insertion sample is visually usable. Do not lower image\_\allowbreak{}realism merely because the scene is stylized, abstract, minimal, semantically unusual, or physically implausible in ways unrelated to the insertion.\par
\noindent\hangindent=1.7em\hangafter=1 - Acceptable context includes minimal abstract geometry, a simple 3D rendering style, unrelated floating objects already present in without, unrelated physically unnatural scene content, and small isolated black rendering artifacts outside the target/insertion region when they do not affect insertion quality or interpretation.\par
\noindent\hangindent=1.7em\hangafter=1 - Accept reasonable contact, support, attachment, and slight embedding between the target and its intended support, such as a cup resting on a coaster or an object attached to a pad/base. Do not call ordinary load-bearing contact, a tight attachment, or shallow support embedding severe mesh penetration. Penalize penetration only when geometry visibly passes through another surface by an implausible amount, produces broken/interleaved faces, or materially damages the target or support silhouette.\par
\noindent\hangindent=1.7em\hangafter=1 - Quality problems include severe visible mesh penetration, broken or missing geometry, abnormal faces, wrong-normal black/bright faces, clearly incorrect material rendering, abnormal exposure or lighting, large black/blank unrendered regions, large unintended solid-color backgrounds, and partial or complete rendering collapse.\par
\noindent\hangindent=1.7em\hangafter=1 - Large-black/low-light score cap: if Image 1 contains a large genuinely unrendered black/blank region, or scene illumination is so low that the target or key scene structure is difficult to recognize, image\_\allowbreak{}realism must be 0, 1, or 2 and never higher. Use 2 when the sample remains usable despite the defect; use 0 or 1 when the blank region or darkness dominates, hides the target, or prevents reliable interpretation. This cap concerns missing content or unusable visibility, not physically plausible black materials or shadows.\par
\noindent\hangindent=1.7em\hangafter=1 - Large unintended solid-color backgrounds of any color cap image\_\allowbreak{}realism at 2 when they visibly replace expected scene content or indicate rendering failure. Apply the stricter flat-gray gate below when the failed region is a substantial near-uniform neutral gray area.\par
\noindent\hangindent=1.7em\hangafter=1 - Mandatory target-integrity gate: first identify the target's complete expected structure, parts, silhouette, and materials from removed\_\allowbreak{}only and the front/right/top views. Then inspect every visible target region in with. The target must appear as a coherent rendered object: expected visible parts must exist, surfaces must be continuous, geometry must not look shattered, torn, erased, hollow, or accidentally clipped, and materials/textures must not be replaced by anomalous featureless black patches. Do not award 5 unless this target-integrity check clearly passes.\par
\noindent\hangindent=1.7em\hangafter=1 - Mandatory target-visibility gate: compare with against removed\_\allowbreak{}only and determine whether a meaningful, recognizable portion of the inserted target is actually visible in the scene image. If the target is almost invisible because it is nearly entirely outside the frame, hidden by scene geometry, reduced to only a tiny sliver or scattered fragments, extremely underexposed, or merged with the background so strongly that its silhouette and material cannot be judged, image\_\allowbreak{}realism must be 0 or 1 and never 2 or higher. Use 0 when the target is effectively absent or cannot be reliably located in with; use 1 when only a tiny insufficient fragment or extremely weak trace remains. As a practical cue, visibility of roughly less than 10 percent of the expected projected target silhouette, or the absence of any coherent identifiable target part, normally triggers this rule. A physically small XS/S object is not a failure when it is still clearly resolved and recognizable.\par
\noindent\hangindent=1.7em\hangafter=1 - Target-defect score caps: a clear but localized target rendering defect, such as a broken surface, missing small part, isolated invalid black patch, wrong-normal patch, texture dropout, or obvious unintended intersection, caps image\_\allowbreak{}realism at 2. A widespread defect, a missing major target part, severe geometry collapse, or a black/corrupt region that materially prevents recognizing or judging the inserted target requires image\_\allowbreak{}realism 0 or 1. A visible but milder suspicious artifact that does not clearly break geometry or materials caps the score at 3.\par
\noindent\hangindent=1.7em\hangafter=1 - For target defects, one direct contradiction with removed\_\allowbreak{}only or the three object views is sufficient evidence; do not require two independent indicators. Examples include an expected target surface becoming a featureless black hole, a target part abruptly disappearing without a foreground occluder, or a material region being replaced by black pixels whose boundary cuts across the target's own geometry. This target-specific rule takes priority over the general black/gray-region rule below.\par
\noindent\hangindent=1.7em\hangafter=1 - Mandatory flat-gray unrendered-region gate: if Image 1 contains a visually substantial region of nearly uniform neutral gray that is clearly empty/unrendered rather than scene geometry, image\_\allowbreak{}realism must be 0 or 1 and never 2 or higher. The decisive evidence is a pure-color-like gray area with almost no texture, material detail, geometry, shading gradient, lighting variation, or physically meaningful structure, often ending at an artificial boundary or replacing an area where scene content should exist. This gate applies even when the gray region does not overlap the target and takes priority over the general non-target two-indicator rule below. Use 0 when the blank gray region dominates or makes the image unusable; otherwise use 1.\par
\noindent\hangindent=1.7em\hangafter=1 - Do not confuse the flat-gray gate with darkness caused by the inserted object or scene lighting. A cast/contact shadow, shaded wall, gray material, fog, or underexposed surface remains valid when it preserves scene texture, geometry, gradual illumination or penumbra, material variation, reflections, or boundaries consistent with physical objects. A dark region is not a pure-gray unrendered region merely because it is large or low-contrast.\par
\noindent\hangindent=1.7em\hangafter=1 - Dark pixels, a large black/gray area, low exposure, or contact with the image boundary are not by themselves rendering failures. Before lowering the score, first identify what geometry occupies that region using with, removed\_\allowbreak{}only, and the three object views. Dark target materials, dark scene objects, fabric folds, physically plausible cast/contact shadows, naturally dark rooms, foreground objects close to the camera, and ordinary image-frame cropping are valid rendered content.\par
\noindent\hangindent=1.7em\hangafter=1 - Evidence of valid rendering includes a coherent object silhouette, continuous geometry, material texture, fabric detail or folds, soft shading gradients, plausible shadow transitions, highlights, and boundaries that follow visible scene geometry. A dark region with any of these cues must not be called flat, blank, or unrendered merely because its average brightness is low.\par
\noindent\hangindent=1.7em\hangafter=1 - General non-target hard failure rule: assign image\_\allowbreak{}realism 0 or 1 for a black/gray/blank scene region only when there is positive visual evidence that rendering content is actually missing or corrupted and it materially prevents judging the target. For non-target/background regions, require at least two of these indicators: (a) the region is genuinely featureless with no geometry, texture, shading gradient, or plausible shadow structure; (b) it has an abrupt boundary that does not follow any object silhouette, occluder, image frame, or physical shadow; (c) it cuts through or replaces target parts that removed\_\allowbreak{}only or the three views show should be visible; (d) it is inconsistent with the corresponding scene geometry in with/without; (e) it produces impossible holes, erased surfaces, or broken topology. This two-indicator requirement does not apply to a directly verified target defect under the target-integrity gate. A dark region appearing only in with is valid only when it is visibly coherent target material or shading, not merely because it overlaps the target location.\par
\noindent\hangindent=1.7em\hangafter=1 - Image-frame cropping is not a failure when enough of the target remains clearly visible and judgeable. However, cropping becomes a hard visibility failure under the target-visibility gate when almost the entire target lies outside the frame or only a tiny unjudgeable fragment remains. Similarly, foreground target parts may legitimately cover the bed, floor, support surface, or other background content.\par
\noindent\hangindent=1.7em\hangafter=1 - Calibration example: dark pillows or other soft furnishings close to the camera may occupy and be cropped by a lower corner. If their silhouettes, fabric folds, shading, and mutual shadows remain coherent, this is valid rendering and must not receive 0 or 1 merely for being dark. It should normally receive 4; use 5 only if the entire image also satisfies the exceptional score-5 gate below.\par
\noindent\hangindent=1.7em\hangafter=1 - The image\_\allowbreak{}realism reason must explicitly report both the target-visibility gate and the target-integrity gate. State whether the target is clearly visible and judgeable in with. If visibility fails, describe what little remains visible and assign 0 or 1. If integrity fails, name the damaged/missing target part or abnormal black region, compare it with removed\_\allowbreak{}only or a specific object view, and state the applied score cap. If both pass, explicitly state that a meaningful target portion is visible and its geometry and materials are coherent. Also state whether the large-black/low-light cap, unintended-solid-color cap, flat-gray gate, or any other general hard failure applies. Do not describe normal dark materials, shadows, foreground coverage, moderate frame cropping, reasonable support contact, attachment, or slight support embedding as corruption or penetration.\par
\noindent\hangindent=1.7em\hangafter=1 - Score 5 is exceptional and must be rare. Award 5 only when the target is clearly visible and fully coherent, placement and contact are physically plausible, geometry/normals/materials/textures are clean, lighting and shadows are consistent, and no visible artifact, suspicious blank area, penetration, corruption, or meaningful quality concern remains anywhere in Image 1. Passing the hard-failure gates alone is not enough for 5. Use 4 for a strong, fully usable image with any minor imperfection, uncertainty, small artifact, slightly awkward integration, or merely ordinary rendering quality. When uncertain between 4 and 5, always choose 4.\par
\noindent 3. Classify object\_\allowbreak{}properties:\par
\noindent\hangindent=1.7em\hangafter=1 - relative\_\allowbreak{}scale: exactly one of "XS", "S", "M", "L", or "XL". Judge the complete removed object group's physical scale relative to the visible scene and nearby objects in Images 1-2. Never use its enlarged crop size in removed\_\allowbreak{}only or the three views.\par
\noindent\hangindent=1.7em\hangafter=1 - XS: an object component or sub-part rather than a normally complete standalone object, such as a handle, knob, bulb, small panel, fixture part, or one component of a larger assembly.\par
\noindent\hangindent=1.7em\hangafter=1 - S: a small standalone or handheld/tabletop object, such as a cup, book, small ornament, small lamp, or compact container.\par
\noindent\hangindent=1.7em\hangafter=1 - M: a normal standalone scene object larger than typical tabletop items but clearly below major furniture scale, such as a floor lamp, medium plant, stool, waste bin, or small appliance.\par
\noindent\hangindent=1.7em\hangafter=1 - L: a large object at furniture or major appliance scale, such as a chair, table, sofa, bed, refrigerator, or substantial cabinet.\par
\noindent\hangindent=1.7em\hangafter=1 - XL: an architectural or near-architectural-scale object occupying a broad scene region, such as a long wall-spanning curtain, a large built-in wardrobe/cabinet wall, a room divider, or another object comparable to a wall opening or building element.\par
\noindent Choose the closest scale class from visible scene evidence. Semantic identity is only an anchor; prioritize the object's actual scale in this scene.\par
\noindent\hangindent=1.7em\hangafter=1 - mechanical\_\allowbreak{}type: exactly "rigid" or "non\_\allowbreak{}rigid". Use "non\_\allowbreak{}rigid" for cloth, curtains, soft furnishings, flexible sheets, deformable vegetation, cables, and similar objects whose shape can deform. Use "rigid" for objects whose shape is normally fixed. Judge the dominant property of a coherent object group.\par
\noindent 4. Classify environment\_\allowbreak{}relations using the exact types and levels below.\par
\noindent Boolean fields:\par
\noindent\hangindent=1.7em\hangafter=1 - has\_\allowbreak{}occlusion: true only when the target is meaningfully partially hidden by foreground scene geometry or participates in a nontrivial layered occlusion boundary. Use removed\_\allowbreak{}only as the primary reference: compare the target silhouette and visible parts in with against the isolated target in removed\_\allowbreak{}only, which represents the same target without scene occluders. If a part is visible in removed\_\allowbreak{}only but disappears or is cut off in with exactly where foreground scene geometry passes in front of it, that is strong evidence of occlusion. Use the front/right/top views as additional references for the target complete structure when needed. Do not classify image-frame cropping, an incomplete removed object, self-occlusion, shadows, or ordinary coverage of the background by a fully visible target as scene occlusion. The reason must identify the foreground occluder and which target part it hides, and should mention the removed\_\allowbreak{}only comparison when it provides the evidence.\par
\noindent\hangindent=1.7em\hangafter=1 - has\_\allowbreak{}strong\_\allowbreak{}perspective\_\allowbreak{}depth: true when the scene has strong depth perspective around the target, such as large near-far scale variation, deep spatial placement, or substantial perspective deformation that makes insertion geometry harder. A normal perspective camera alone is not enough.\par
\noindent\hangindent=1.7em\hangafter=1 - has\_\allowbreak{}reflective\_\allowbreak{}surface\_\allowbreak{}interaction: this field is specifically about the inserted target changing the appearance of a pre-existing reflective or transmissive scene surface. Use Image 7 only to locate scene regions changed by inserting the target, then inspect the same coordinates in the original with and without images. Image 7 colors mean: blue = the scene became darker, red = the scene became brighter, green = the scene mainly changed color. The color itself does not determine the label.\par
\noindent Return true only when all of the following are satisfied: (1) the changed region is outside the inserted target; (2) the receiving scene surface is visibly identifiable as glossy, polished, metallic, glass, water, or another genuinely reflective/refractive material; (3) the aligned with/without comparison shows a clear target-caused change on that surface; (4) the change has reflection, specular-highlight, or refraction characteristics rather than ordinary diffuse illumination; and (5) the change is readily visible to a human at normal viewing scale. Examples that may qualify include the target or its color appearing in a glossy cabinet or polished floor, a target-caused specular highlight changing on metal or glass, or visible refraction through a transmissive scene surface.\par
\noindent Return false for ordinary diffuse brightening or darkening of walls, ceilings, fabric, matte wood, or other non-reflective surfaces; generic bounce light or color bleeding; illumination cast by an emissive target; target cast/contact shadows; ordinary light blocking; ambient occlusion; and any vague change for which the receiving surface cannot be confidently identified as reflective or transmissive. A recognizable mirror reflection belongs exclusively to has\_\allowbreak{}mirror\_\allowbreak{}reflection\_\allowbreak{}interaction and must be false here.\par
\noindent Material identity alone is not enough: the reflective/refractive content or highlight on that surface must visibly change between with and without. Conversely, a changed region alone is not enough: do not infer reflection merely from red, green, or blue pixels in Image 7. Image 7 is amplified and is never sufficient evidence by itself. When the material, cause, or human visibility is uncertain, return false.\par
\noindent The reason must name the receiving scene surface and its visible reflective/transmissive material cue, identify the changed reflection/highlight/refraction, state whether Image 7 shows blue/red/green there, and confirm that the change remains readily visible in the original with/without images. For false, explicitly state which required condition is missing, such as no identifiable reflective receiver, only diffuse illumination, only shadow/light blocking, only a mirror reflection, or insufficient human-visible change.\par
\noindent\hangindent=1.7em\hangafter=1 - has\_\allowbreak{}mirror\_\allowbreak{}reflection\_\allowbreak{}interaction: true only when one of the following two evidence paths is satisfied:\par
\noindent A. Scene-mirror change: a mirror or mirror-like scene surface contains recognizable reflected content that changes because of the target insertion. Compare the same mirror region at aligned coordinates in with and without. The reflected target, or another recognizable reflected scene object/structure whose mirror image is altered or occluded by the target, must appear, disappear, move, or visibly change between the two images. A mirror merely being present is false. An unchanged mirror image is false, even if it contains many recognizable scene objects.\par
\noindent B. Target-is-mirror: the inserted target itself is a mirror or contains a genuine mirror surface, and in with that target mirror visibly reflects recognizable objects or spatial structure from the surrounding environment. Use removed\_\allowbreak{}only and the three views to confirm that the reflective surface belongs to the target, then identify the specific reflected environment content visible after insertion. This path does not require a with/without comparison inside the target region because the target mirror is absent in without.\par
\noindent Mandatory decision procedure: first determine whether the candidate mirror belongs to the pre-existing scene or to the inserted target. For a pre-existing scene mirror, true requires a concrete aligned with/without change in its recognizable reflected content caused by the target. For a target mirror, true requires recognizable environment content inside the target mirror. Otherwise return false.\par
\noindent Do not use this field for ordinary glossy highlights, reflected color, illumination changes, vague silhouettes, blurry metallic reflections, dark screens, glass without recognizable reflected content, or changes outside the mirror region. Those may belong to has\_\allowbreak{}reflective\_\allowbreak{}surface\_\allowbreak{}interaction but are not mirror interaction. The reason must name the mirror owner, the reflected object or structure, and either the exact with/without mirror-content change for path A or the recognizable environment content reflected by the target mirror for path B.\par
\noindent\hangindent=1.7em\hangafter=1 - has\_\allowbreak{}emissive\_\allowbreak{}large\_\allowbreak{}scale\_\allowbreak{}lighting\_\allowbreak{}change: true only when the target is self-emissive and its presence causes a visible large-scale illumination or color change in the surrounding environment. A glowing pixel or bright lamp surface without substantial environmental lighting change is false.\par
\smallskip
\noindent Three-level integer fields (must be JSON integers 0, 1, or 2, not booleans):\par
\noindent\hangindent=1.7em\hangafter=1 - has\_\allowbreak{}mechanical\_\allowbreak{}interaction:\par
\noindent\hangindent=1.7em\hangafter=1 - 0 (none): no clearly visible support, contact, suspension, attachment, or other meaningful force relation with scene geometry can be identified.\par
\noindent\hangindent=1.7em\hangafter=1 - 1 (simple): ordinary static support and gravity balance through one simple support arrangement, such as an object resting on a table, shelf, floor, or a single regular base. A target simply mounted on or placed against a wall also counts as simple when no explicit load-bearing suspension mechanism or positional constraint is visible; an object that could plausibly occupy an arbitrary wall position must not be promoted to level 2 merely because it is described as hanging. Several feet touching the same flat floor still normally count as simple unless they create a genuinely nontrivial balance.\par
\noindent\hangindent=1.7em\hangafter=1 - 2 (complex): nontrivial force balance involving multiple independent support/contact points or constraints from different directions, tension, clamping, leaning with multiple force-bearing contacts, or another mechanically coupled arrangement whose stability depends on more than ordinary single-surface support. A hanging or suspended target qualifies as level 2 only when the images show an explicit suspension dependency and load path, such as a visible hook, rope, cable, chain, bracket, slot, ceiling attachment, or multiple suspension points that constrain and support the target. Do not infer complex suspension from semantic identity or wall placement alone.\par
\noindent Mere proximity without visible force-bearing contact is 0. Assign level 2 only from visible mechanical evidence, not from an assumption about how the object is normally installed. State the visible support/contact points, suspension mechanism, and force relation in the reason.\par
\noindent\hangindent=1.7em\hangafter=1 - has\_\allowbreak{}complex\_\allowbreak{}lighting\_\allowbreak{}shadows:\par
\noindent\hangindent=1.7em\hangafter=1 - 0 (none): no clearly identifiable target-related shadow is visible; extremely weak generic ambient occlusion that cannot be confidently attributed to the target also counts as 0.\par
\noindent\hangindent=1.7em\hangafter=1 - 1 (simple): exactly one dominant target-related contact/cast shadow direction or one coherent shadow lobe is visible under otherwise simple lighting. A soft shadow still counts, but level 1 is allowed only after checking all sides of the target and finding no second independently directed target shadow.\par
\noindent\hangindent=1.7em\hangafter=1 - 2 (complex): assign level 2 whenever two or more distinct target-related cast-shadow lobes or projections are visible in different directions, on opposite sides of the target, or on different receiving surfaces. This is direct evidence of multiple or spatially varied light contributions. The shadows may be soft, faint, partially overlapping, or unequal in strength; they do not need to be crisp, high-contrast, intricate, or produced by obvious visible lamps. Also use level 2 for split illumination across the target, multiple light directions, overlapping shadows, strongly spatially varying light, or intricate/high-contrast shadow patterns.\par
\noindent Mandatory shadow inspection procedure:\par
\noindent 1. Locate the complete target using with and removed\_\allowbreak{}only, then inspect the floor/support and the full perimeter on its left, right, front, and back image sides rather than stopping at the strongest shadow.\par
\noindent 2. Compare the same coordinates in with and without. A dark or shaded lobe that appears with the target and disappears or weakens without it is target-related evidence. Use Image 7 to locate subtle non-target changes, then verify them in with/without.\par
\noindent 3. Inspect shadows cast by nearby scene objects as supporting evidence for the scene light directions. If those shadows reveal illumination from multiple directions, actively search for the corresponding weaker target projections; scene-object shadows are clues, but level 2 still requires visible target-related shadow or split-light evidence.\par
\noindent 4. If two target shadows extend to the left and right, or otherwise diverge in direction, classify level 2 even when one is much softer or less visible than the other. A common level-2 case is one strong lobe on one side plus one broad faint lobe on the opposite side. Do not merge them into one soft shadow and do not reject the weaker lobe merely because its boundary is diffuse.\par
\noindent 5. Distinguish ambient occlusion from a weak cast shadow spatially. Ambient occlusion is restricted to a narrow contact rim or crevice that tightly hugs the target base with no directional outward extent. A dark region that spreads outward from the silhouette over the receiving surface, has a displaced or directional lobe shape, or occupies a broader area than the immediate contact rim is a shadow projection, even if it is faint and soft.\par
\noindent 6. Use scene lighting evidence as a tie-breaker. When nearby objects show shadows or highlights consistent with two light directions, and the target has one strong lobe plus a weaker outward-spreading lobe in the corresponding second direction, count both target lobes and assign level 2. Do not call the weaker lobe ambient occlusion unless it is strictly confined to the narrow target-contact boundary.\par
\noindent Calibration example: a round chair on a rug with a visible projection to its right and a weaker but spatially extended projection to its left is level 2, not level 1. Softness and unequal intensity do not reduce the number of shadow directions.\par
\noindent Do not count truly generic ambient occlusion or texture darkening as an additional cast shadow. The reason must enumerate every detected target shadow by image-relative direction and receiving surface. A level-1 reason must explicitly state that every other dark region is confined to a narrow contact rim with no outward directional extent; saying only that it is faint or soft is not sufficient to reject it.\par
\par\medskip\noindent\textbf{[Output requirements]}\par\nobreak
\noindent Return only one JSON object wrapped in a json code fence. Use exactly the English keys shown below. Every classification and every relation must have a concise, evidence-based reason.\par
\noindent Keep every reason to exactly one sentence of at most 20 English words. Keep the complete response below 700 English words; do not add commentary outside the JSON.\par
\smallskip
\begin{Verbatim}[fontsize=\fontsize{8.3}{10}\selectfont,breaklines=true,breakanywhere=true,breaksymbolleft={},breaksymbolright={}]
{
  "insertion_description": "insert a table lamp on the side table",
  "image_realism": 4,
  "object_properties": {
    "relative_scale": "S",
    "mechanical_type": "rigid"
  },
  "environment_relations": {
    "has_occlusion": false,
    "has_strong_perspective_depth": false,
    "has_mechanical_interaction": 1,
    "has_complex_lighting_shadows": 1,
    "has_reflective_surface_interaction": false,
    "has_mirror_reflection_interaction": false,
    "has_emissive_large_scale_lighting_change": true
  },
  "reason": {
    "image_realism": "The scene is coherent and has no severe rendering defects.",
    "relative_scale": "The lamp is a small tabletop object relative to the room and nearby furniture.",
    "mechanical_type": "The lamp has a fixed rigid body.",
    "environment_relations": {
      "has_occlusion": "The complete lamp is visible without meaningful foreground occlusion.",
      "has_strong_perspective_depth": "The lamp is viewed in a locally shallow region without strong near-far deformation.",
      "has_mechanical_interaction": "Level 1: the lamp has a simple single-surface support relation with the tabletop.",
      "has_complex_lighting_shadows": "Level 1: one straightforward cast shadow from the lamp is visible on the tabletop.",
      "has_reflective_surface_interaction": "No nearby reflective surface visibly changes because of the lamp.",
      "has_mirror_reflection_interaction": "The lamp is not visibly reflected by a scene mirror and does not itself contain a mirror reflecting recognizable scene objects.",
      "has_emissive_large_scale_lighting_change": "The lamp visibly changes illumination over a broad surrounding region."
    }
  }
}
\end{Verbatim}
\end{tcolorbox}
\endgroup

\clearpage
\section{User Study Protocol}
\label{app:user-study-protocol}

We collect complete responses from 27 distinct participants on 25 in-the-wild insertion cases, comparing PhysInsert, Insert-Anything, DIRECT, and Nano Banana Pro. Method identities are hidden, and case and method orders are randomized. These cases form a separate user-study set constructed by combining scenes from RealBokeh with foreground objects from Distinctions-646. With the background and object reference visible, participants rate side-by-side results from 1 to 5 (higher is better) on lighting integration (including object illumination and cast shadows), perspective, pose, identity preservation, and background preservation. We average over cases within each participant, then across participants; the overall score equally weights the five criteria.

The error bars in Figure~\ref{fig:user-study} show 95\% percentile confidence intervals estimated from 50,000 participant-level bootstrap resamples, with the 25 cases held fixed.

\end{document}